\documentclass[10pt,twocolumn,letterpaper]{article}

\usepackage{cvpr}              

\usepackage{graphicx}
\usepackage{amsmath}
\usepackage{amssymb}
\usepackage{booktabs}
\usepackage{bm}
\usepackage{mathtools}
\usepackage{balance}
\usepackage{multicol}

\usepackage[pagebackref,breaklinks,colorlinks]{hyperref}

\usepackage[capitalize]{cleveref}
\crefname{section}{Sec.}{Secs.}
\Crefname{section}{Section}{Sections}
\Crefname{table}{Table}{Tables}
\crefname{table}{Tab.}{Tabs.}

\usepackage[flushleft]{threeparttable}

\newcommand{\tablestyle}[2]{\setlength{\tabcolsep}{#1}\renewcommand{\arraystretch}{#2}\centering\footnotesize}
\newlength\savewidth

\newtheorem{property}{Property}
\newtheorem{theorem}{Theorem}
\newtheorem{lemma}[theorem]{Lemma}

\newtheorem{theorem1}{Theorem}
\newenvironment{proof}{{\noindent\bf Proof.}\quad}{\hfill $\square$\par}

\usepackage{diagbox}
\usepackage{makecell}
\usepackage{colortbl}
\usepackage{wrapfig}

\usepackage{amsmath,amsfonts,bm}

\DeclarePairedDelimiterX{\lin}[2]{\langle}{\rangle}{#1, #2}
\DeclarePairedDelimiterX{\abs}[1]{\lvert}{\rvert}{#1}
\DeclarePairedDelimiterX{\norm}[1]{\lVert}{\rVert}{#1}
\DeclarePairedDelimiterX{\cbr}[1]{\{}{\}}{#1} 
\DeclarePairedDelimiterX{\rbr}[1]{(}{)}{#1} 
\DeclarePairedDelimiterX{\sbr}[1]{[}{]}{#1} 

  \providecommand{\R}{\mathbb{R}} 

  \DeclareMathOperator{\expect}{\mathbb{E}}

\def\eqref#1{equation~\ref{#1}}

\def\1{\bm{1}}

\def\rvg{{\mathbf{g}}}

\def\vu{{\bm{u}}}

\def\vw{{\bm{w}}}

\def\m1{{\bm{1}}}

\def\mB{{\bm{B}}}

\def\mP{{\bm{P}}}

\def\mU{{\bm{U}}}
\def\mV{{\bm{V}}}
\def\mW{{\bm{W}}}
\def\mX{{\bm{X}}}

\DeclareMathAlphabet{\mathsfit}{\encodingdefault}{\sfdefault}{m}{sl}
\SetMathAlphabet{\mathsfit}{bold}{\encodingdefault}{\sfdefault}{bx}{n}

\def\gB{{\mathcal{B}}}

\def\gG{{\mathcal{G}}}

\def\gM{{\mathcal{M}}}

\def\gS{{\mathcal{S}}}
\def\gT{{\mathcal{T}}}

\def\sB{{\mathbb{B}}}

\def\sP{{\mathbb{P}}}

\def\sU{{\mathbb{U}}}

\DeclareMathOperator*{\argmax}{arg\,max}
\DeclareMathOperator*{\argmin}{arg\,min}

\usepackage{graphicx}
\usepackage{float}

\newfloat{figtab}{htb}{fgtb}
\makeatletter
  \newcommand\figcaption{\def\@captype{figure}\caption}
  \newcommand\tabcaption{\def\@captype{table}\caption}
\makeatother

\usepackage[dvipsnames]{xcolor}
\hypersetup{
    colorlinks,
    citecolor=RoyalBlue,
}

\definecolor{Gray}{gray}{0.5}

\definecolor{Highlight}{HTML}{39b54a}
\definecolor{Modify}{HTML}{2240F0}

\definecolor{Highlight}{HTML}{3B7F27}
\newcommand{\gap}[1]{
	\fontsize{6pt}{1em}\selectfont{\textcolor{Highlight}{\hskip-0.1em\bf{({#1}{$\times$})}}}
}

\newcommand{\bgap}[1]{
	\fontsize{6pt}{1em}\selectfont{\textcolor{Highlight}{\hskip-0.1em({#1}{$\times$})}}
}

\newcommand{\ggap}[1]{
	\fontsize{6pt}{1em}\selectfont{\textcolor{Gray}{\hskip-0.1em({#1}{$\times$})}}
}

\usepackage{arydshln}
\usepackage{multirow}
\usepackage{makecell}

\def\cvprPaperID{6151} 
\def\confName{CVPR}
\def\confYear{2023}

\newcommand*{\affaddr}[1]{#1} 
\newcommand*{\affmark}[1][*]{\textsuperscript{#1}}

\makeatletter
\newcommand{\printfnsymbol}[1]{\textsuperscript{\@fnsymbol{#1}}}
\makeatother
\makeatletter
\def\thanks#1{\protected@xdef\@thanks{\@thanks
        \protect\footnotetext{#1}}}
\makeatother

\begin{document}

\title{Compacting Binary Neural Networks by Sparse Kernel Selection}

\author{
	Yikai Wang\affmark[1]\quad Wenbing Huang\affmark[2]\quad Yinpeng Dong\affmark[1,3]\quad Fuchun Sun\affmark[1]\quad\vspace{0.05in}Anbang Yao\affmark[4]\\
	\affaddr{\affmark[1]BNRist Center, State Key Lab on Intelligent Technology and Systems,\\ Department of Computer Science and Technology, Tsinghua University} \\\affaddr{\affmark[2]Gaoling School of Artificial Intelligence, Renmin University of China}\quad
	\affaddr{\affmark[3]RealAI}\quad
	\affaddr{\affmark[4]Intel Labs China}\\
	\tt\small{\{yikaiw,fcsun,dongyinpeng\}@tsinghua.edu.cn,hwenbing@126.com,anbang.yao@intel.com}\\
}

\maketitle

\begin{abstract}
Binary Neural Network (BNN) represents convolution weights with 1-bit  values, which enhances the efficiency of storage and computation. This paper is motivated by a previously revealed phenomenon that the binary kernels in successful BNNs are nearly power-law distributed: their values are mostly clustered into a small number of codewords. This phenomenon encourages us to compact typical BNNs and obtain further close performance through learning non-repetitive kernels within a binary kernel subspace. Specifically, we regard the binarization process as kernel grouping in terms of a binary codebook, and our task lies in learning to select a smaller subset of codewords from the full codebook. We then leverage the Gumbel-Sinkhorn technique to approximate the codeword selection process, and develop the Permutation Straight-Through Estimator (PSTE) that is able to not only optimize the selection process end-to-end but also maintain the non-repetitive occupancy of selected codewords. Experiments verify that our method reduces both the model size and bit-wise computational costs, and achieves accuracy improvements compared with state-of-the-art BNNs under comparable budgets.
\end{abstract}

\section{Introduction}
\label{sec:intro}
It is crucial to design compact Deep Neural Networks (DNNs) which allow the model deployment on resource-constrained embedded devices, since most powerful DNNs including ResNets~\cite{he2016deep} and DenseNets~\cite{cvprHuangLMW17} are storage costly with deep and rich building blocks piled up. Plenty of approaches have been proposed to compress DNNs, among which network quantization~\cite{corrdorefa,jmlrHubaraCSEB17,cvprZhuangSTL018} is able to reduce memory footprints as well as accelerate the inference speed by converting full-precision weights to discrete values. 
Binary Neural Networks (BNNs)~\cite{corrBengioLC13,nipsHubaraCSEB16} belong to the family of network quantization but they further constrict the parameter representations to binary values ($\pm1$). In this way, the model is largely compressed. More importantly, floating-point additions and multiplications in conventional DNNs are less required and mostly reduced to bit-wise operations that are well supported by fast inference accelerators~\cite{eccvRastegariORF16}, particularly when activations are binarized as well.
To some extent, this makes BNNs more computationally efficient than other  compression techniques, \emph{e.g.}, network pruning~\cite{neuripsHanPTD15,iccvHeZS17,iccvLiuLSHYZ17} and switchable models~\cite{DBLP:journals/corr/abs-1903-05134,DBLP:conf/iclr/YuYXYH19,wang2020rsnets}.

\begin{figure*}[t]
\centering
\hskip-0.2em
\includegraphics[scale=0.335]{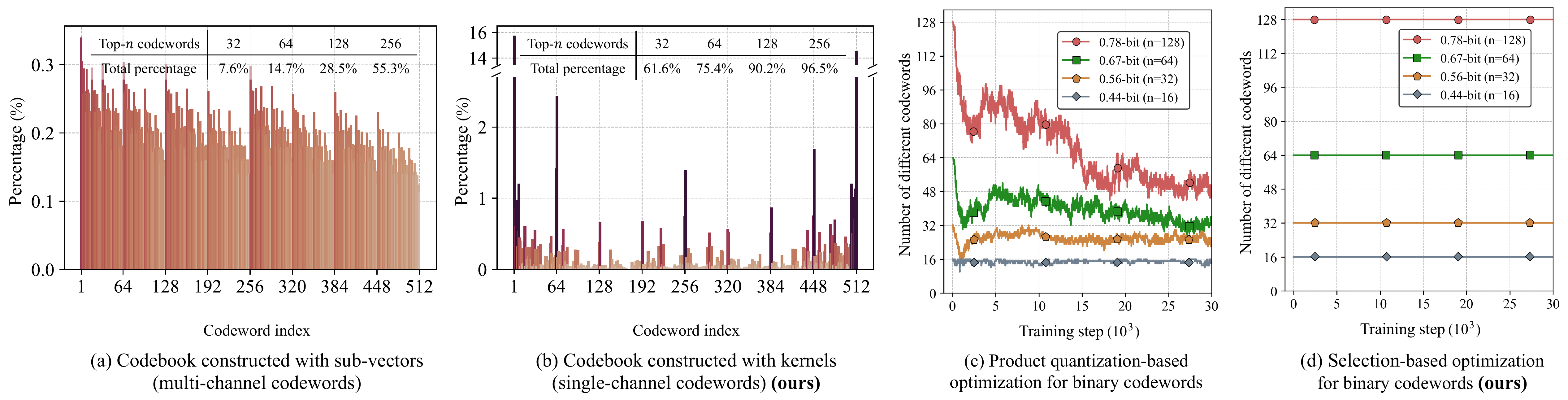}
\vskip-0.5em
\caption{Codebook distributions under different decomposition approaches. Statistics in both sub-figures are collected from the same BNN model (XNOR-Net~\cite{eccvRastegariORF16} upon ResNet-18~\cite{he2016deep}) well-trained on ImageNet~\cite{Deng2009ImageNet}. In (a), each codeword is a flattened sub-vector (of size $1\times9$). In (b), each codeword is a $3\times3$ convolution kernel. The codebook in either sub-figure consists of $2^9=512$ different codewords.  Upper tables provide the total percentages of the top-$n$ most frequent codewords. In (c), we observe that the sub-codebook highly degenerates during training, since codewords tend to be repetitive when being updated independently. While in (d), the diversity of codewords preserves, which implies the superiority of our selection-based learning.}
\label{pic:distribution}
\vskip-0.5em
\end{figure*}

Whilst a variety of methods are proposed to improve the performance of BNNs,  seldom is there a focus on discussing how the learnt binary kernels are distributed in BNNs. A recent work SNN~\cite{wang2021snn} demonstrates that, by choosing typical convolutional BNN models~\cite{eccvRastegariORF16,cvprQinGLSWYS20,DBLP:conf/eccv/LiuSSC20} well trained on ImageNet and displaying the distribution of the $3\times3$ kernels along all possible $2^{3\times3}$ binary values (\emph{a.k.a.} codewords),   these kernels nearly obey the power-law distribution: only a small portion of codewords are activated for the most time. Such a phenomenon is re-illustrated in Figure~\ref{pic:distribution}(b). This observation motivates SNN to restrict the size of the codebook by removing those hardly-selected codewords. As a result, SNN is able to compact BNN further since indexing the kernels with a smaller size of codebook results in a compression ratio of $\log_2(n)/\log_2(N)$, where $n$ and $N$ are separately the sizes of the compact and full codebooks.

However, given that the size of codebook is limited (only $512$), the sub-codebook degenerates during training since codewords are likely to become repetitive. Therefore, we believe the clustering property of kernels can be further exploited during the training of BNNs. To do so, we reformulate the binary quantization process as a grouping task that selects, for each kernel, the nearest codeword from a binary sub-codebook which is obtained by selecting optimal codewords from the full one. To pursue an optimal solution and retain the non-repetitive occupancy of the selected codewords, we first convert the sub-codebook selection problem to a permutation learning task. However, learning the permutation matrix is non-differential since the permutation matrix is valued with only 0/1 entries. Inspired by the idea in~\cite{iclrMenaBLS18}, we introduce the Gumbel-Sinkhorn operation to generate a continuous and differential approximation of the permutation matrix. During training, we further develop Permutation Straight-Through Estimator (PSTE), a novel method that tunes the approximated permutation matrix end-to-end while maintaining the binary property of the selected codewords. The details are provided in \textsection~\ref{sec:sinkhorn} and \textsection~\ref{sec:pste}. 
We further provide the complexity analysis in \textsection~\ref{sec:complexity}.

Extensive results on  image classification and object detection demonstrate that our architecture noticeably reduces the model size as well as the computational burden. For example, by representing ResNet-18 with 0.56-bit per weight on ImageNet, our method brings in 214$\times$ saving of bit-wise operations, and 58$\times$ reduction of the model size. Though state-of-the-art BNNs have achieved remarkable compression efficiency, we believe that further compacting BNNs is still beneficial, by which we can adopt deeper, wider, and thus more expressive architectures without exceeding the complexity budget than BNNs. For example, our  0.56-bit ResNet-34 obtains 1.7\% higher top-1 accuracy than the state-of-the-art BNN on ResNet-18, while its computational costs are lower and the storage costs are almost the same.       

Existing methods~\cite{DBLP:journals/corr/abs-2010-02778,NeurIPSLeeKKJPY20} (apart from SNN~\cite{wang2021snn}) that also attempt to obtain more compact models than BNNs are quite different with ours as will be described in \textsection~\ref{sec:related}. One of the crucial points is that their codewords are sub-vectors from (flattened) convolution weights across multiple channels, whereas our each codeword corresponds to a complete kernel that maintains the spatial dimensions (weight and height) of a single channel. The reason why we formulate the codebook in this way stems from the observation in Figure~\ref{pic:distribution}(b), where the kernels are sparsely clustered. Differently, as shown in Figure~\ref{pic:distribution}(a), the codewords are nearly uniformly activated if the codebook is constructed from flattened sub-vectors, which could because the patterns of the input are spatially selective but channel-wise uniformly distributed. It is hence potential that our method may recover better expressivity of BNNs by following this natural characteristic. In addition, we optimize the codewords via non-repetitive selection from a fixed codebook, which rigorously ensures the dissimilarity between every two codewords and thus enables more capacity than the product quantization method used in~\cite{DBLP:journals/corr/abs-2010-02778}, as compared in Figure~\ref{pic:distribution}(c)(d).
On ImageNet with the same  backbone, our method exceeds \cite{DBLP:journals/corr/abs-2010-02778} and \cite{NeurIPSLeeKKJPY20} by $6.6\%$ and $4.5\%$ top-1 accuracies, respectively.

\section{Related Work}
\label{sec:related}

\textbf{BNNs}. Network quantization methods~\cite{corrdorefa,jmlrHubaraCSEB17,DBLP:journals/ijcv/DongNLCSZ19,cvprZhuangSTL018} convert network weights to low-bit values and are appealing for resource-limited devices given the superiority in efficiency. As an extreme solution of quantization, BNNs~\cite{corrBengioLC13,nipsHubaraCSEB16,eccvRastegariORF16,DBLP:conf/eccv/TuCRW22} represent weights and activations with 1-bit ($\pm1$) values, bringing 32$\times$ storage compression ratio and 58$\times$ practical computational reduction on CPU as reported by~\cite{eccvRastegariORF16}. BNNs usually adopt a non-differentiable sign function during the forward pass and the Straight-Through Estimator (STE)~\cite{corrBengioLC13} for gradient back-propagation.
Many attempts are proposed to narrow the performance gap between BNNs and their real-valued counterparts. XNOR-Net~\cite{eccvRastegariORF16} adopts floating-point parameters as scaling factors to reduce the quantization error. Bi-Real~\cite{ijcvLiuLWYLC20} proposes to add ResNet-like shortcuts to reduce the information loss during binarization. ABC-Net~\cite{neuripsABCNet} linearly combines multiple binary weight bases to further approximate full-precision weights. ReActNet~\cite{DBLP:conf/eccv/LiuSSC20} generalizes activation functions to capture the distribution reshape and shift. New architectures for BNNs can be searched~\cite{DBLP:conf/iccvw/ShenHXW19}  or designed~\cite{DBLP:journals/corr/abs-2010-03558} to further improve the trade-off between performance and efficiency.

\textbf{Compacting BNNs.}
Our work focuses on an orthogonal venue and investigates how to compact BNNs further. 
Previously, SNN~\cite{wang2021snn} reveals that binary kernels learnt at convolutional layers of a BNN model are likely to be distributed over kernel subsets. Based on this, SNN randomly samples layer-specific  binary kernel subsets and refines them during training. However, the  optimization of SNN is easy to attain repetitive binary kernels, \emph{i.e.}, degenerated  subsets,    leading to a noticeable performance drop compared with conventional BNN. Another method sharing a similar motivation with us is the fractional quantization
(FleXOR)~\cite{NeurIPSLeeKKJPY20}, which encrypts the sub-vectors of flattened weights to low-dimensional binary codes. Yet, FleXOR cannot track which weights share the same encrypted code, and thus it is needed to decrypt the compressed model back to the full BNN for inference. In our method, the reconstruction of the corresponding BNN is unnecessary, since the computation can be realized in the space of codewords, leading to further reduction of bit-wise computations as detailed in \textsection~\ref{sec:complexity}.  
Another research close to our paper is 
SLBF~\cite{DBLP:journals/corr/abs-2010-02778} that applies the idea of stacking low-dimensional  filters~\cite{DBLP:conf/icml/YangWLCXS0X19} and the product quantization~\cite{pamiJegouDS11,pamiGeHK014,iclrStockJGGJ20} to BNNs. Similar to~\cite{NeurIPSLeeKKJPY20}, this method
splits the (flattened) weights into sub-vectors as codewords along the channel direction. As already demonstrated in \textsection~\ref{sec:intro}, our method leverages kernel-wise codebook creation by selection-based codeword optimization, yielding much lower quantization errors than~\cite{NeurIPSLeeKKJPY20,DBLP:journals/corr/abs-2010-02778}.

\section{Sparse Kernel Selection}

In this section, we introduce how to compact and accelerate BNN further by \underline{Spar}se \underline{K}ernel \underline{S}election, abbreviated as \textbf{Sparks}. Towards this goal, we first formulate the quantization process as grouping convolution kernels into a certain binary codebook. We then show that a more compact sub-codebook can be learnt end-to-end via Gumbel-Sinkhorn ranking. To enable the optimization of the ranking while keeping the binary property, we further propose the Permutation Straight-Through Estimator (PSTE) technique with the convergence analysis. Finally, we contrast the complexity of the model with BNN, and demonstrate that our method is able to not only compress BNN further but also accelerate the speed of BNN during inference. 

\subsection{Binarization below 1-Bit}
\label{sec:preliminaries}
Prior to going further, we first provide the notations and necessary preliminaries used in our paper. Let $\mW\in \R^{C_{\text{out}}\times C_{\text{in}}\times K\times K}$ be the convolution weights, where $C_{\text{out}}$ and $C_{\text{in}}$ are the numbers of the output and input channels, respectively, and $K$ is the kernel size. As discussed in \textsection~\ref{sec:intro}, we are interested in the quantization for each specific kernel, denoted with a lowercase letter as $\vw\in\R^{K\times K}$. The quantization process aims to map full-precision weights to a smaller set of discrete finite values.
Specifically, we conduct the function $\mathrm{sign}$ for each kernel, resulting in $\hat{\vw}=\mathrm{sign}(\vw)\in\sB$ where $\sB=\{-1,+1\}^{K\times K}$ is the set consisting of 1-bit per weight. In what follows, $\sB$ is called the (full) codebook, and each element in $\sB$ is a codeword.

Generally, the quantization can be rewritten as an optimization problem $\hat{\vw}=\argmin_{\vu\in\sB}\|\vu-\vw\|_2$ that grouping each kernel $\vw$ to its nearest codeword in  $\sB$, where $\|\cdot\|_2$ denotes the $\ell_2$ norm. We state this equivalence in the form below, and the proof is provided in Appendix.

 \begin{property}
We denote $\sB=\{-1,+1\}^{K\times K}$ as the codebook of binary kernels. For each $\vw\in\R^{K\times K}$, the binary kernel $\hat{\vw}$ can be derived by a grouping process:
\begin{equation}
\label{eq:grouping}
  \hat{\vw}= \mathrm{sign}(\vw)=\argmin_{\vu\in\sB}\|\vu-\vw\|_2.
\end{equation}
\end{property}
\vskip-0.5em

Since the codebook size $|\sB|=2^{K\times K}$, the memory complexity of BNN is equal to $K\times K$. Given Equation~\ref{eq:grouping}, one may want to know if we can further reduce the complexity of BNN by, for example, sampling a smaller subset of the codebook $\sB$ to replace $\sB$ in Equation~\ref{eq:grouping}. This is also motivated by Figure~\ref{pic:distribution}(b) where the learnt kernels of BNNs are sparsely clustered into a small number of codewords. In this way, each kernel is represented below $K^2$-bits and thus averagely, each weight is represented less than 1-bit. We thus recast the grouping as
\begin{eqnarray}
\label{eq:grouping-re}
\hat{\vw}=\argmin_{\vu\in\sU}\|\vu-\vw\|_2, \text{ \emph{s.t.} } \sU\subseteq\sB.
\end{eqnarray}
We denote $|\sU|=n$ and $|\sB|=N$. By Equation~\ref{eq:grouping-re}, each binary kernel occupies $\log_2(n)$ bits as it can be represented by an index in $\{1,2,\cdots,n\}$. Thus we obtain a compression ratio of $\log_2(n)/\log_2(N)$.

Different choice of the sub-codebook $\sU$ from $\sB$ potentially delivers different performance. How can we determine the proper selection we prefer? One possible solution is to optimize codewords and kernels simultaneously by making use of the product quantization method~\cite{iclrStockJGGJ20,DBLP:journals/corr/abs-2010-02778}. Nevertheless, this method updates each codeword independently and is prone to deriving repetitive codewords if the optimization space is constrained (as values are limited to $\pm1$). As a consequence, it will hinder the diversity of the codebook and limit the expressivity of the model, which will be verified in \textsection~\ref{sec:ablation}. Another straightforward method would be sampling the most frequent $n$ codewords from a learnt target BNN (depicted in Figure~\ref{pic:distribution}(b)). Yet, such a solution is suboptimal since it solely depends on the weight distribution of the eventual model without the involvement of the specific training dynamics. In the following subsections, we will propose to tackle the sub-codebook selection via an end-to-end approach while retaining the non-repeatability of the codewords.

\vspace{-0.1em}
\subsection{Sub-Codebook Selection via Permutation}
\label{sec:sinkhorn}

We  learn a permutation of  codewords in $\sB$ according to their effects on the target loss function, so that the selection of the first $n$ codewords is able to optimize the final performance of the target task.  The designed permutation learning keeps the binary property of the selected codewords. 

For convenience, we index  codewords in $\sB$ as a matrix column by column, formulated as   $\mB=[\vu_1; \vu_2; \cdots;\vu_{N}]\in\{\pm1\}^{K^2\times N}$, where each codeword in $\sB$ is flattened to a $K^2$-dimensional vector. Similarly, we convert $\sU$ as $\mU=[\vu_{s_1}; \vu_{s_2}; \cdots,\vu_{s_{n}}]\in\{\pm1\}^{K^2\times n}$, where $s_i\in\{1,2,\cdots,N\}$ is the index of the $i$-th selected codeword. We denote the selection matrix as $\mV\in\{0,1\}^{N\times n}$, then 
\begin{eqnarray}
\label{eq:selection}
\;\mU=\mB\mV,
\end{eqnarray}
where the entries of $\mV$ satisfy $\mV_{s_i,i}=1$ for $i=1,\cdots,n$ and are zeros otherwise. The selection by Equation~\ref{eq:selection} is permutation-dependent; in other words, if we permute the element of $\sB$, we may obtain different $\sU$. Hence, how to select $\sU$ becomes how to first permute $\sB$ and then output $\sU$ by Equation~\ref{eq:selection}. 
We denote $\sP_N$ the set of $N$-dimensional permutation matrices: $\sP_N = \{\mP\in \{0,1\}^{N\times N}\, |\, \mP \m1_N =\m1_N, \mP^\top \m1_N=\m1_N\}$, where $\m1_N$ is an $N$-dimensional column vector of ones. The optimization problem in Equation~\ref{eq:grouping-re} is transformed into
\begin{eqnarray}
\label{eq:permutation}
\hat{\vw}=\argmin_{\vu\in\sU} \|\vu-\vw\|_2, \text{ \emph{s.t.} } \mU=\mB\mP\mV, \mP\in\sP_N,
\end{eqnarray}
where $\mV$ is fixed as a certain initial selection. 

Now, the goal is how to determine a proper permutation matrix $\mP$. Basically, we can design a neural network to output $\mP$, and then embed it into the off-the-shelf CNN for the downstream task. 
Unfortunately, this pipeline fails as the permutation matrix $\mP$ is discrete, whose values are occupied with 0 or 1, making the permutation network non-differentiable. Joining the recent advancement of permutation learning, we leverage the method proposed by~\cite{corr-abs-1106-1925} that approximates the permutation matrix by its continuous and differentiable relaxation---the Sinkhorn operator~\cite{sinkhorn1964relationship}.

Given a matrix $\mX\in\R^{N\times N} (N=|\sB|)$, the Sinkhorn operator over $\gS(\mX)$ is proceeded as follow,
\begin{align}
\label{eq:sinkhorn}
\gS^{0}(\mX) &= \exp(\mX),\\
\gS^{k}(\mX) &= \gT_c\left(\gT_r(S^{k-1}(\mX))\right),\\
\gS(\mX) &=\lim_{k\rightarrow\infty} \gS^{k}(\mX),
\end{align}
where $\gT_r(\mX)= \mX \oslash (\mX \m1_N\m1_N^\top)$ and $ \gT_c(\mX)=\mX \oslash (\m1_N\m1_N^\top\mX $) are the row-wise and column-wise normalization operators, and $\oslash$ denotes the element-wise division. For stability purpose, both normalization operators are calculated in the log domain in practice. The work by~\cite{sinkhorn1964relationship} proved that $\gS(\mX)$ belongs to the Birkhoff polytope---the set of doubly stochastic matrices.

Through adding a temperature $\tau$, it can be proved that $\lim_{\tau\rightarrow 0^{+}}\gS(\mX/\tau)=\argmax_{\mP\in\sP_N}\|\mP-\mX\|_{2}$ holds almost surely~\cite{iclrMenaBLS18}. It means we obtain an approximated permutation matrix $\gS^k(\mX)$ (that is closest to $\mX$) with sufficiently large $k$ and small $\tau$. Inspired by~\cite{iclrJangGP17}, we also add a Gumbel noise to make the result  follow the Gumbel-Matching distribution $\gG.\gM.(\mX)$, namely, $\gS^k((\mX+\epsilon)/\tau)$, where $\epsilon$ is sampled from standard i.i.d. Gumbel distribution. 

By substituting the Gumbel-Sinkhorn matrix into Equation~\ref{eq:selection}, we characterize the sub-codebook selection as
\begin{eqnarray}
\label{eq:selection-final}
\;\mU =\,\mB\gS^k((\mX+\epsilon)/\tau)\mV,
\end{eqnarray}
where $\mV$ is fixed as a certain initial selection as mentioned, $\mX$ is regarded as a learnable parameter, $k$ and $\tau$ are hyper-parameters. For $\mV$, we can simply let the entries to be zeros unless $\mV_{i,i}=1$, where $i=1,\cdots, n$, which indicates selecting the first $n$ columns from $\mB\gS^k((\mX+\epsilon)/\tau)$.

\begin{figure*}[t]
\centering
\vskip-0.3em
\includegraphics[scale=0.24]{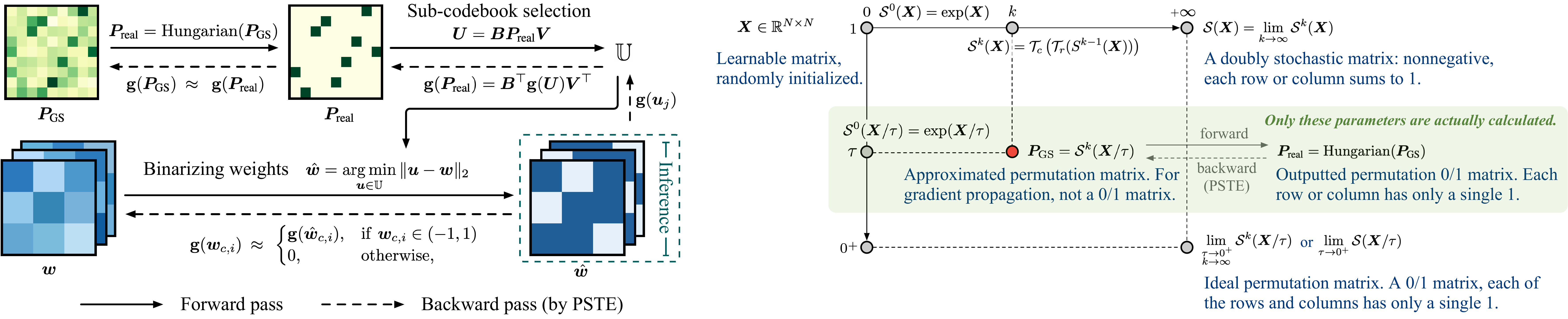}
\caption{\textbf{Left:} A schematic overview of the optimization process. Full-precision weights are binarized by grouping to the nearest codeword of the sub-codebook $\sU$, which is obtained by Gumbel-Sinkhorn and optimized by PSTE. Forward and backward passes illustrate how the network calculates and updates. \textbf{Right}: Relationship of notations for a better understanding. The horizontal axis and vertical axis stand for the values of $k$ and $\tau$, respectively. Only parameters in the green region are actually calculated during training, while the others are only for the understanding purpose. All these parameters are NOT needed during inference.}
\label{pic:framework}
\vskip-0.6em
\end{figure*}

\vspace{-0.2em}
\subsection{Learning by PSTE}
\label{sec:pste}
Recalling that both $k$ and $\tau$ are finitely valued, the Gumbel-Sinkhorn matrix $\mP_{\mathrm{GS}}=\gS^k((\mX+\epsilon)/\tau)$ is not strictly a permutation matrix with 0/1 entries. This will violate the binary property of $\mU$ by Equation~\ref{eq:selection-final}, making the binarization of Equation~\ref{eq:grouping-re} meaningless. 
To address this issue, we derive the exact permutation matrix $\mP_{\mathrm{real}}$ of $\mP_{\mathrm{GS}}$ by making use of the Hungarian algorithm~\cite{munkres1957algorithms} during the forward pass. By treating the $\mP_{\mathrm{GS}}$ as a reward matrix, deriving $\mP_{\mathrm{real}}$ becomes an assignment problem that can be solved by the Hungarian method in polynomial time. We summarize the forward update of the convolution kernel $\vw_c\in\R^{K^2}$ for each input and output channel as follow,
\begin{align}
\mP_{\mathrm{real}} &= \mathrm{Hungarian}(\mP_{\mathrm{GS}}),\\
\mU &= \mB\mP_{\mathrm{real}}\mV,\\
\hat{\vw}_c &= \argmin_{\vu\in\sU}\|\vu-\vw_c\|_{2},
\end{align}
where $\mathrm{Hungarian}(\cdot)$ denotes the Hungarian algorithm. 
\vspace{-0.2em}

In the backward pass, we transfer the gradient of the exact permutation matrix directly to the Gumbel-Sinkhorn matrix. This is inspired by the Straight-Through Estimator (STE) technique~\cite{corrBengioLC13} in previous literature. 
We call our method PSTE for its specification to permutation learning here. The backward pass is depicted below,
\begin{eqnarray}
\label{eq:ste}
\rvg({\vw_{c,i}}) \hskip-0.5em&\approx&\hskip-0.3em 
\begin{cases}
\rvg({\hat{\vw}_{c,i}}) ,& \mathrm{if} \,\, \vw_{c,i} \in \left(-1, 1\right),\\
0,& \mathrm{otherwise},
\end{cases}\\
\label{eq:dictioanry-gradient}
\rvg({\vu_j})  \hskip-0.5em&=& \hskip-0.8em \sum_{c=1}^{C_{\text{in}}\times C_{\text{out}}}\hskip-0.1em\rvg({\hat{\vw}_c})\cdot\mathbb{I}_{\vu_j=\argmin_{\vu\in\sU}\|\vu-\vw_c\|_{2}},\\
\rvg({\mP_{\mathrm{real}}}) \hskip-0.5em&=&\hskip-0.2em  \mB^{\top}\rvg({\mU})\mV^{\top},\\
\rvg({\mP_{\mathrm{GS}}}) \hskip-0.5em&\approx& \hskip-0.2em \rvg({\mP_{\mathrm{real}}}),\label{eq:pgs}
\end{eqnarray}
where $\rvg(\cdot)$ computes the gradient. $\vw_{c,i}$ and $\hat{\vw}_{c,i}$ denote the $i$-th entries of $\vw_c$ and $\hat{\vw}_c$, respectively, with $i=1,2,\cdots,K^2$. $\mathbb{I}_{\{\cdot\}}$ defines the indicator function. Particularly, Equation~\ref{eq:ste} follows the idea of STE and Equation~\ref{eq:dictioanry-gradient} assigns the gradient of the binary weight to its nearest codeword. In practice, all forward and backward passes can be implemented by matrix/tensor operations, and thus our method is computationally friendly on GPUs. 

An overall framework including the forward and backward processes is illustrated in Figure~\ref{pic:framework}.

\textbf{Convergence analysis.} 
Besides using STE to  update  $\vw$  in Equation~\ref{eq:ste}, we approximate the gradient of the Gumbel-Sinkhorn matrix $\mP_{\mathrm{GS}}$ with $\mP_{\mathrm{real}}$ in Equation~\ref{eq:pgs}, which, inevitably, will cause variations in training dynamics. Fortunately, we have the following theorem to guarantee the convergence for  sufficiently large $k$ and small $\tau$.

\def\PGS{\mP_{\mathrm{GS}}}
\def\Pr{\mP_{\mathrm{real}}}

\begin{lemma}
\label{lemma1}
For sufficiently large k and small $\tau$, we define the entropy of a doubly-stochastic matrix $\mP$ as $h(\mP)=-\sum_{i,j}P_{i,j}\log P_{i,j}$, and denote the rate of convergence for the Sinkhorn operator as $r\,(0<r<1)$\footnote{The Sinkhorn operator has a rate of convergence $r$ bounded by a value lower than 1 as proved by~\cite{journalssiammaxKnight08}.}. There exists a convergence series $s_\tau$ ($s_\tau\rightarrow0$ when $\tau\rightarrow0^+$) that satisfies  
\begin{align}
\|\Pr-\PGS\|_2^2 = \mathcal{O} \big(s_\tau^2+  r^{2k}\big).
\end{align}
\end{lemma}

\begin{theorem1}\label{thm:nonconvex}  
Assume that the training objective $f$ \emph{w.r.t.}$\,\,\mP_{\mathrm{GS}}$ is $L$-smooth, and the stochastic gradient of $\mP_{\mathrm{real}}$ is bounded by $\mathbb{E}\|\mathbf{g}(\mP_{\mathrm{real}})\|_2^2\leq \sigma^2$. Denote the rate of convergence for the Sinkhorn operator as $r\,(0<r<1)$ and the stationary point as $\mP_{\mathrm{GS}}^{\star}$. Let the learning rate of PSTE be $\eta = \frac{c}{\sqrt{T}}$ with $c=\sqrt{\frac{f(\mP_{\mathrm{GS}}^0)-f(\mP_{\mathrm{GS}}^\star)}{{L \sigma^2}}}$. For a uniformly chosen $\bm{u}$ from the iterates $\{\Pr^0,\cdots,\Pr^T\}$, concretely $\bm{u}=\Pr^t$ with the probability $p_t=\frac{1}{T+1}$, it holds in expectation over the stochasticity and the selection of $\bm{u}$ :
\begin{small}
\begin{equation}
\mathbb{E}\|\nabla f(\bm{u})\|_2^2 = \mathcal{O} \left( \sigma\sqrt{\frac{f(\PGS^0)-f(\PGS^\star)}{T/L}}  + L^2\big(s_\tau^2 + r^{2k}\big) \right)\,.\vspace{-2mm}
\end{equation}\end{small}
\end{theorem1}

Note that in Theorem~\ref{thm:nonconvex}, the objective function $f$ could be a non-convex function, which accords with the case when using a neural network. Proofs for Lemma~\ref{lemma1} and Theorem~\ref{thm:nonconvex} are provided in Appendix.

\subsection{Complexity Analysis During Inference}
\label{sec:complexity}

\textbf{Storage}. We consider the convolutional layer with $3\times 3$ kernels. In a conventional binary convolutional layer, the weights requires $C_{\text{out}}\times C_{\text{in}}\times K\times K$ bits. 
For our method, we only  store the sub-codebook $\sU$ and the index of each kernel by Equation~\ref{eq:grouping-re}. Storing $\sU$ needs $n\times K\times K$ bits, where $n=|\sU|$. Since $n\leq N=2^{K\times K}\ll C_{\text{out}}\times C_{\text{in}}$ for many popular CNNs (\emph{e.g.}, ResNets~\cite{he2016deep}), particularly if all layers share the same $\sU$ which is the case of our implementation, only the indexing process counts majorly. The indexing process needs $\log_2(n)$ bits for a kernel, hence indexing all kernels takes $C_{\text{out}}\times C_{\text{in}}\times \log_2(n)$. As a result, the ratio of storage saving by our method is $\log_2(n)/(K\times K)=\log_2(n)/\log_2(N)\leq1$  compared to a conventional BNN.

\begin{table*}[t]
\begin{center}
\begin{small}
\hskip -0.07in
\tablestyle{1pt}{1.1}
\resizebox{0.8\linewidth}{!}{
	\begin{tabular}{lccccp{0.1cm}lcccc}
		\toprule
		{Method}&Bit-width& Accuracy &Storage&BOPs&&{Method}&Bit-width& Accuracy  &Storage&BOPs\\
		\bf{(ResNet-18)}&(W$/$A) & Top-1$\,$(\%)& (Mbit)&($\times10^{8}$)&&\bf{(VGG-small)}&(W$/$A) &Top-1$\,$(\%)& (Mbit)&($\times10^{8}$)\\
		\midrule
		Full-precision &32$/$32&94.8 & 351.5\ggap{1}&350.3\ggap{1} && Full-precision &32$/$32&94.1 & 146.2\ggap{1}&386.6\ggap{1}\\
		\cmidrule(r){1-11}
		XNOR-Net~\cite{eccvRastegariORF16}&1$/$1&90.2 & 11.0\ggap{32}&5.47\ggap{64}&& XNOR-Net~\cite{eccvRastegariORF16}&1$/$1&89.8&4.57\ggap{32}&6.03\ggap{64}\\
		Bi-RealNet~\cite{ijcvLiuLWYLC20}&1$/$1&90.2 & 11.0\ggap{32}&5.47\ggap{64}&&LAB~\cite{Loss-Aware-BNN}&1$/$1&87.7 & 4.57\ggap{32}&6.03\ggap{64}\\
		RAD~\cite{Regularize-act-distribution}&1$/$1&90.5 & 11.0\ggap{32}&5.47\ggap{64}&&RAD~\cite{Regularize-act-distribution}&1$/$1&90.0 &4.57\ggap{32}&6.03\ggap{64}\\
		IR-Net~\cite{cvprQinGLSWYS20}&1$/$1&91.5& 11.0\ggap{32}&5.47\ggap{64}&&IR-Net~\cite{cvprQinGLSWYS20}&1$/$1&90.4&4.57\ggap{32}&6.03\ggap{64}\\
		RBNN~\cite{DBLP:conf/nips/LinJX00WHL20}&1$/$1&92.2& 11.0\ggap{32}&5.47\ggap{64}&&RBNN~\cite{DBLP:conf/nips/LinJX00WHL20}&1$/$1&91.3&4.57\ggap{32}&6.03\ggap{64}\\
		ReActNet~\cite{DBLP:conf/eccv/LiuSSC20}&1$/$1&92.3& 11.0\ggap{32}&5.47\ggap{64}&&SLB~\cite{DBLP:conf/nips/YangWHX0T020}&1$/$1&92.0&4.57\ggap{32}&6.03\ggap{64}\\
		\cmidrule(r){1-11}
		SLBF~\cite{DBLP:journals/corr/abs-2010-02778}& 0.55$/$1 & 89.3\scriptsize{$\,\pm$0.5} & 6.05\bgap{58}&$\;\,$2.94\bgap{119} && SLBF~\cite{DBLP:journals/corr/abs-2010-02778} & 0.53$/$1 &89.4\scriptsize{$\,\pm$0.4}  & 2.42\bgap{60}&$\;$3.17\bgap{122}\\
		FleXOR~\cite{NeurIPSLeeKKJPY20} & 0.80$/$1 &90.9\scriptsize{$\,\pm$0.2}  & 8.80\bgap{40}&5.47\ggap{64}&&FleXOR~\cite{NeurIPSLeeKKJPY20} & 0.80$/$1 &90.6\scriptsize{$\,\pm$0.1}  & 3.66\bgap{40}&6.03\ggap{64}\\
		FleXOR~\cite{NeurIPSLeeKKJPY20} & 0.60$/$1 &89.8\scriptsize{$\,\pm$0.3}  & 6.60\bgap{53}&5.47\ggap{64}&&FleXOR~\cite{NeurIPSLeeKKJPY20} & 0.60$/$1 &89.2\scriptsize{$\,\pm$0.2}  & 2.74\bgap{53}&6.03\ggap{64}\\
\cmidrule(r){1-11}
		Sparks (ours) & 0.78$/$1 &92.2\scriptsize{$\,\pm$0.1}  &  8.57\gap{41}&3.96\gap{88} && Sparks (ours) & 0.78$/$1 &91.7\scriptsize{$\,\pm$0.2}  & 3.55\gap{41}&$\;$3.46\gap{112}\\
		Sparks (ours) & 0.67$/$1 &92.0\scriptsize{$\,\pm$0.2}  &  7.32\gap{48}&$\;$2.97\gap{118} && Sparks (ours) & 0.67$/$1 &91.6\scriptsize{$\,\pm$0.1}  & 3.05\gap{48}&$\;$1.94\gap{199}\\
		Sparks (ours) & 0.56$/$1 & 91.5\scriptsize{$\,\pm$0.3} & 6.10\gap{58}&$\;$1.63\gap{215} && Sparks (ours) & 0.56$/$1 &91.3\scriptsize{$\,\pm$0.3}  & 2.54\gap{58}&$\;$1.13\gap{342}\\
		Sparks (ours) & 0.44$/$1 & 90.8\scriptsize{$\,\pm$0.2} & 4.88\gap{72}&$\,$0.97\gap{361}&&Sparks (ours) & 0.44$/$1 &90.8\scriptsize{$\,\pm$0.3}  & 2.03\gap{72}&$\;$0.74\gap{522}\\
		\bottomrule
	\end{tabular}}
\caption{Comparisons of top-1 accuracies with state-of-the-art methods on the CIFAR10 dataset. }\label{table:cifar10}
\end{small}
\end{center}
\vskip -0.2in
\end{table*}

\textbf{Computation}.
We adopt BOPs to measure the computational costs and follow the calculation method in~\cite{DBLP:conf/iclr/MartinezYBT20,ijcvLiuLWYLC20,DBLP:conf/eccv/LiuSSC20,eccvRastegariORF16} where each BOP represents two bit-wise operations.
In a conventional BNN, the convolution operation between the input feature maps ($C_{\text{in}}\times H\times W$) and weights ($C_{\text{out}}\times C_{\text{in}}\times K\times K$) takes $ \mathrm{BOPs}=H\times W\times C_{\text{in}}\times K^2\times C_{\text{out}}, $
where $H$ and $W$ are the height and width of the feature map, respectively.
For our method, the kernel grouping process implies that some weights will share the same value, which enables us to further reduce $\mathrm{BOPs}$.
To do so, we pre-calculate the convolutions between the input feature maps ($C_{\text{in}}\times H\times W$)
and each codeword ($K\times K$). For facility, we reshape the codeword as $1\times C_{\text{in}}\times K\times K$ by repeating the second dimension over $C_{\text{in}}$ times. Then, the pre-convolution for all codewords gives rise to a tensor $\gT$ ($n \times C_{\text{in}}\times H\times W$) and costs $\mathrm{BOPs}_1=H\times W\times C_{\text{in}}\times K^2\times n.$ We  reconstruct the convolution result between the input feature maps and the convolution weights given the pre-calculated tensor $\gT$. Specifically, for each output channel, we query  indices of the input channels \emph{w.r.t.} $\sU$, collect  feature maps from $\gT$ according to the indices, and sum the feature maps as the final result. This process consumes $\mathrm{BOPs}_2=C_{\text{out}}\times (C_{\text{in}}\times H\times W-1)/2$. Therefore, Sparks needs $\mathrm{BOPs}_1+\mathrm{BOPs}_2$ which is far less than $\mathrm{BOPs}$ when $K=3$, as $ n < C_{out}$ in general.

\section{Experiments}
\label{sec:exp}
Our method is evaluated on two tasks: image classification and object detection (in Appendix).  For image classification, we contrast the performance of our Sparks with state-of-the-art (SOTA) methods on CIFAR10~\cite{krizhevsky2009learning} and ImageNet (ILSVRC2012)~\cite{Deng2009ImageNet}  following the standard data splits. 

\textbf{Implementation.}\label{sec:implementation}
We follow the standard binarization in ReActNet~\cite{DBLP:conf/eccv/LiuSSC20} and perform a two-stage training. First, the network is trained from scratch with binarized activations and real-valued weights. Second, the network takes the weights from the first step and both weights and activations are binarized. As suggested by~\cite{eccvRastegariORF16,ijcvLiuLWYLC20}, we keep the weights and activations in the first convolutional and the last fully-connected layers to be real-valued. More implementation details (\emph{e.g.}, learning rate, epoch) are in the Appendix.

We speed up our training twice by exploiting the symmetry of the binary kernels: for each codeword in the codebook $|\sB|$, its ``opposite'' term (with opposite signs) is also contained in $|\sB|$. Speed-up details are contained in  Appendix.  Hyper-parameters are $k=10$  and $\tau=10^{-2}$ by default. 

On ImageNet,  Sparks needs 30.2 hours to train ResNet-18 based on 8 V100s, and the BNN baseline~\cite{DBLP:conf/eccv/LiuSSC20} needs 24.5 hours. The computation overhead is acceptable. Training Sparks can be easily  accelerated as presented in Appendix

Calculations of storage and BOPs savings are based on the measurements used in~\cite{eccvRastegariORF16,ijcvLiuLWYLC20}. Specifically, compared to full-precision convolutional layers with 32-bit weights, using 1-bit weights and activations gains up to a $\sim$32$\times$ storage saving; in addition, the convolution operation could be implemented by the bit-wise xnor operation followed by a popcount operation, which leads to a $\sim$64$\times$ computational saving. Throughout our results, we provide the amount of bit-wise parameters in all binarized layers as the storage. 

\vspace{-0.05in}

\subsection{Comparisons with SOTA Methods}
\label{sec:classification_results}

\vspace{-0.02in}

\begin{table*}[t]
\begin{center}
\begin{small}
\tablestyle{10pt}{0.9}
\vskip-0.5em
\resizebox{0.8\linewidth}{!}{
	\begin{tabular}{p{2.7cm}p{1.6cm}<{\centering}p{0.8cm}<{\centering}p{0.8cm}<{\centering}p{1.8cm}<{\centering}p{1.6cm}<{\centering}}
		\toprule
		\multirow{2}*{Method}&Bit-width& \multicolumn{2}{c}{Accuracy (\%)} &Storage&BOPs\\
		&(W$/$A) & Top-1& Top-5 & (Mbit)&($\times10^{9}$)\\
		\midrule
		Full-precision&32$/$32&69.6&89.2 &351.5&107.2\ggap{1}\\
		\cmidrule(r){1-6}
		BNN~\cite{nipsHubaraCSEB16}&1$/$1&42.2 &69.2& 11.0\ggap{32}&1.70\ggap{63}\\
		XNOR-Net~\cite{eccvRastegariORF16}&1$/$1&51.2 &73.2&  11.0\ggap{32}&1.70\ggap{63}\\
		Bi-RealNet~\cite{ijcvLiuLWYLC20}&1$/$1&56.4 &79.5&11.0\ggap{32}&1.68\ggap{64}\\
		IR-Net~\cite{cvprQinGLSWYS20}&1$/$1&58.1&80.0&  11.0\ggap{32}&1.68\ggap{64}\\
		LNS~\cite{DBLP:conf/icml/HanWXXWX20}&1$/$1&59.4&81.7&  11.0\ggap{32}&1.68\ggap{64}\\
		RBNN~\cite{DBLP:conf/nips/LinJX00WHL20}&1$/$1&59.9&81.9&11.0\ggap{32}&1.68\ggap{64}\\
		Ensemble-BNN~\cite{DBLP:conf/cvpr/ZhuDS19}& (1$/$1)$\times6$&61.0&-&$\,$65.9\ggap{5.3}&10.6\ggap{10}\\
		ABC-Net~\cite{neuripsABCNet}& $\;\,$(1$/$1)$\times5^2$&65.0&85.9&274.5\ggap{1.3}&$\,$42.5\ggap{2.5}\\
		Real-to-Bin~\cite{DBLP:conf/iclr/MartinezYBT20}&1$/$1&65.4&86.2&  11.0\ggap{32}&1.68\ggap{64}\\
		ReActNet~\cite{DBLP:conf/eccv/LiuSSC20}&1$/$1&65.9&86.4& 11.0\ggap{32}&1.68\ggap{64}\\
		\cmidrule(r){1-6}
		SLBF~\cite{DBLP:journals/corr/abs-2010-02778}& 0.55$/$1 & 57.7 &80.2& 6.05\bgap{58}&$\;$0.92\bgap{117}\\
		SLBF~\cite{DBLP:journals/corr/abs-2010-02778}& 0.31$/$1 & 52.5 &76.1& $\;$3.41\bgap{103}&$\;$0.98\bgap{110}\\
		FleXOR~\cite{NeurIPSLeeKKJPY20} & 0.80$/$1 &62.4 &83.0& 8.80\bgap{40}&1.68\ggap{64}\\
		FleXOR~\cite{NeurIPSLeeKKJPY20} & 0.60$/$1 &59.8  &81.9& 6.60\bgap{53}&1.68\ggap{64}\\
		\cmidrule(r){1-6}
		Sparks (ours) &0.78$/$1&65.5& 86.2&8.57\gap{41}&1.22\gap{88}\\
		Sparks (ours) &0.67$/$1&65.0& 86.0&7.32\gap{48}&$\;$0.88\gap{122}\\
		Sparks (ours) &0.56$/$1&64.3& 85.6&6.10\gap{58}&$\;$0.50\gap{214}\\		\bottomrule
	\end{tabular}}
\vskip -0.05in
\caption{Comparisons of top-1 and top-5 accuracies with state-of-the-art methods on ImageNet based on ResNet-18. Calculation details for the storage and BOPs are provided in Appendix.}
\label{table:imagenet}
\end{small}
\end{center}
\end{table*}

\begin{table*}[t]
\vskip -0.1in

\begin{center}
\begin{small}
\tablestyle{6pt}{0.9}
\resizebox{0.8\linewidth}{!}{
	\begin{tabular}{p{2.2cm}p{2.4cm}<{\centering}p{1.9cm}<{\centering}p{0.8cm}<{\centering}p{0.8cm}<{\centering}p{1.4cm}<{\centering}p{1cm}<{\centering}}
		\toprule
		\multirow{2}*{Method}&\multirow{2}*{Backbone}&Bit-width& \multicolumn{2}{c}{Accuracy (\%)} &Storage&BOPs\\
		&&(W$/$A)& Top-1& Top-5 &(Mbit)&($\times10^{9}$)\\
		\midrule		
		ReActNet~\cite{DBLP:conf/eccv/LiuSSC20}&ResNet-18&1$/$1&65.9&86.4&11.0&1.68\\
		Sparks-wide&\makecell{ResNet-18\\(+ABC-Net~\cite{neuripsABCNet})} &(0.56$/$1)$\times3$&\textbf{66.7}& \textbf{86.9}&18.3&\textbf{1.50}\\
		Sparks-deep &ResNet-34& 0.56$/$1&\textbf{67.6}& \textbf{87.5}&11.7&\textbf{0.96}\\
		Sparks-deep &ResNet-34& 0.44$/$1&\textbf{66.4}& \textbf{86.7}&\textbf{9.4}&\textbf{0.58}\\
		\bottomrule
	\end{tabular}}

\end{small}
\end{center}
\vskip-0.16in
\caption{Results when extending our Sparks to wider or deeper models.}
\label{table:wide_deep}
\vskip -0.05in
\end{table*}

\begin{figure*}[t!]
\centering

\includegraphics[scale=0.495]{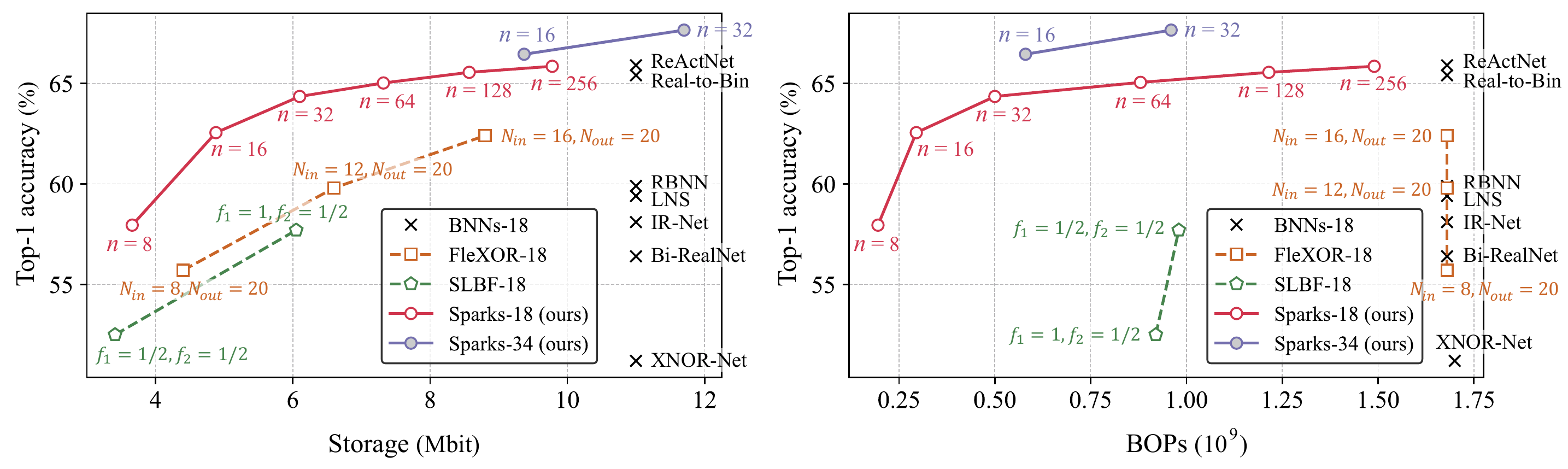}
\vskip-0.5em\vskip-0.1em
\figcaption{Trade-off between performance and complexity on ImageNet. For all methods, -18 indicates using ResNet-18 as the backbone, and -34 indicates ResNet-34. The symbol $n$ is the sub-codebook size as defined in~\textsection~\ref{sec:preliminaries}; $N_{in},N_{out},f_1,f_2$ are hyper-parameters that control the complexity as defined in FleXOR~\cite{NeurIPSLeeKKJPY20} and SLBF~\cite{DBLP:journals/corr/abs-2010-02778}.}
\label{pic:compare_sotas}
\vskip-0.6em
\vskip -0.02in
\end{figure*}

\textbf{Evaluation on CIFAR10.} 
Table \ref{table:cifar10} provides the performance comparisons with SOTA BNNs on CIFAR10. Moreover, SLBF~\cite{DBLP:journals/corr/abs-2010-02778} and 
FleXOR~\cite{NeurIPSLeeKKJPY20} that derive weights below 1-bit are re-implemented upon the same backbone (ReActNet) and same binarization settings (both weights and activations are binarized) as our method for a fair comparison.
By setting $n$ to 32, 64, and 128, we obtain networks of 0.56-bit, 0.67-bit, and 0.78-bit, respectively. 
Clearly, our approach is able to achieve accuracies close to standard BNNs with a much lower cost of storage and BOPs on both ResNet-18 and VGG-small. 
FleXOR also compacts BNNs but delivers no BOPs reduction. SLBF reduces both storage and BOPs but suffers from more accuracy drops as it conducts convolution in the space of multi-channel codewords, leading to much larger quantization errors, as previously described in Figure~\ref{pic:distribution}. 
Our Sparks remarkably outperforms SLBF and FleXOR under almost the same compression level of storage (\emph{e.g.} for ResNet-18, our 0.56-bit gains accuracy 91.5\%, while  0.60-bit FleXOR and 0.55-bit SLBF yield 89.8\% and 89.3\%, respectively), which verify the effectiveness of our proposed method.

\begin{figure*}[t]
\centering
\vskip-0.5em
\hskip-0.3em\includegraphics[scale=0.51]{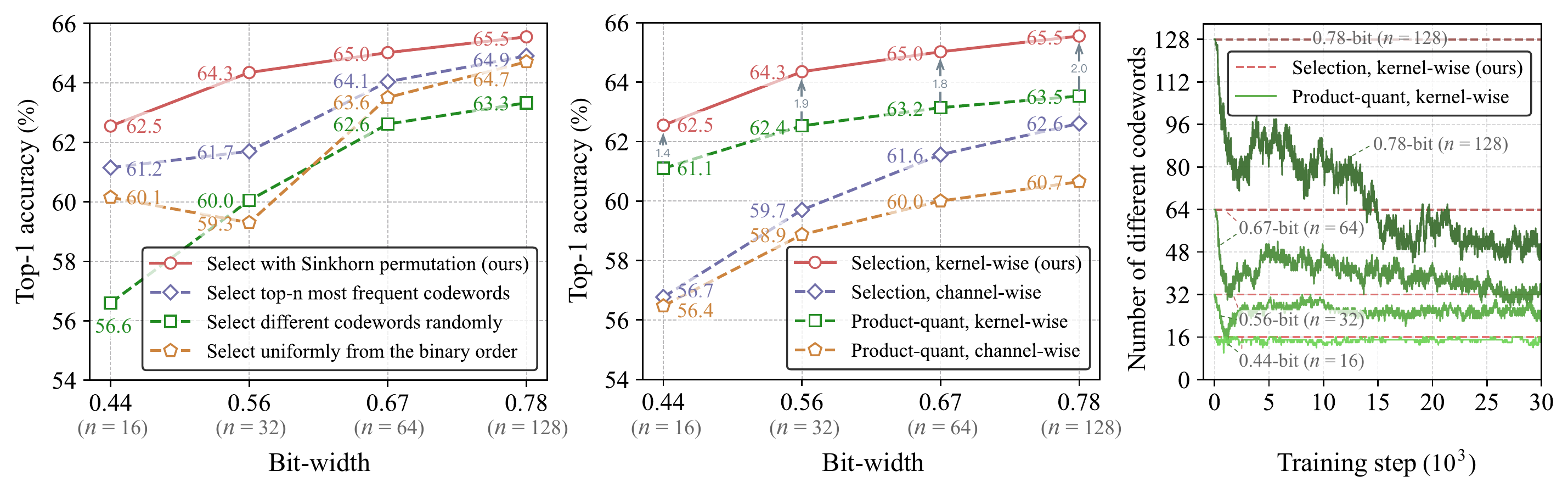}
\vskip-0.5em
\figcaption{Ablation studies on ImageNet with ResNet-18. All experiments adopt the same baseline for a fair comparison. \textbf{Left}: comparisons of different methods to select codewords. \textbf{Middle:} kernel-wise vs channel-wise codewords, and selection-based vs product-quantization(quant)-based learning. Note that ``Product-quant, kernel-wise'' refers to our implementation of SNN~\cite{wang2021snn} under a fair two-stage training framework for improved performance. \textbf{Right:}  sub-codebook degeneration when learning codewords with product-quantization.}
\label{pic:ablation_study}
\vskip-0.3em
\end{figure*}

\begin{figure}[t!]
\centering
\hskip-0.6em
\vskip0.15em
\includegraphics[scale=0.47]{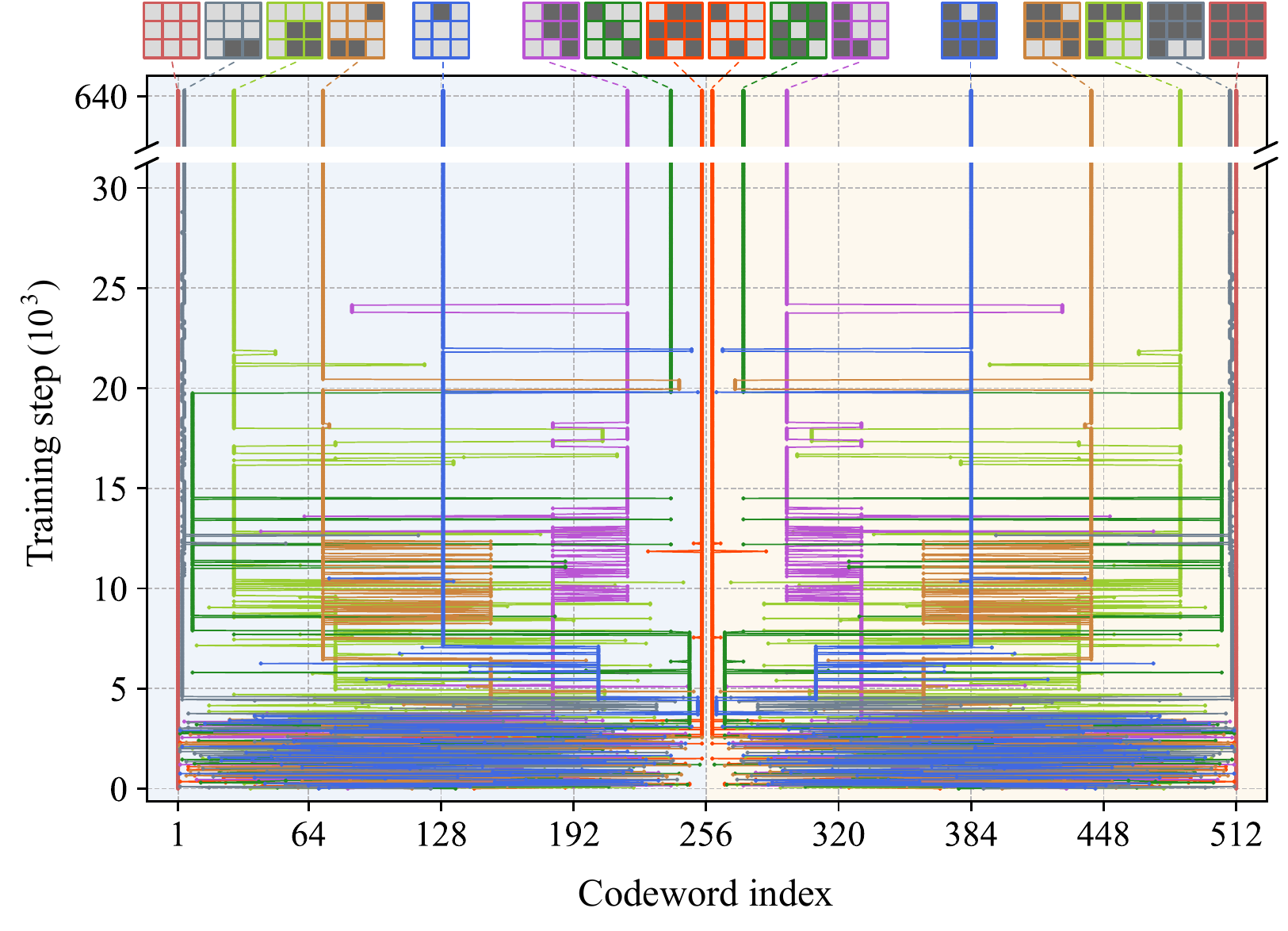}
\vskip-0.5em
\figcaption{Codewords selection during training where $n=16$. We have visualized the final selected 16 codewords at the top of the figure, where light gray indicates $-1$ and dark gray indicates $+1$.}
\label{pic:patterns_step}
\vskip-0.6em
\vskip -0.02in
\vskip-0.5em
\end{figure}

\textbf{Evaluation on ImageNet.} 
In Table \ref{table:imagenet}, we compare Sparks with SOTA methods upon ResNet-18 on ImageNet.
Consistent with the results on CIFAR10,  our method is able to achieve competitive classification accuracy compared to SOTA BNNs, with a dramatic drop in model size and computation time.  
Under comparable bit-widths, our 0.56-bit method exceeds 0.55-bit SLBF and 0.6-bit FleXOR by $6.6\%$ and $4.5\%$, respectively, with even fewer BOPs. Thanks to the remarkable benefit in model compression, we can apply our Sparks on wider and deeper models while staying the same complexity budget as BNNs. For example, in Table~\ref{table:wide_deep}, we follow ABC-Net~\cite{neuripsABCNet} and apply our 0.56-bit model on ResNet-18 by using three branches of convolutions and a single branch of activation, denoted as Sparks-wide; we also adopt 0.56-bit and 0.44-bit Sparks on ResNet-34 (almost twice in depth compared to ResNet-18), both denoted as Sparks-deep. We observe that all our variants surpass ReActNet-18 (currently the best ResNet-18-based BNN) with almost the same or even less cost in model complexity. Specifically, our 0.44-bit Sparks-deep defeats ReActNet-18 in accuracy with the least cost of complexity. To better visualize the trade-off between performance and efficiency, we contrast our models against existing methods with varying storage and BOPs in Figure~\ref{pic:compare_sotas}. We leave the comparison with SNN~\cite{wang2021snn} in Figure~\ref{pic:ablation_study}, since we re-implemented SNN under a fair two-stage  pipeline, which improves the absolute accuracy of SNN by $6.9\%\sim8.1\%$.

\subsection{Ablation Studies}
\label{sec:ablation}

\vspace{-0.1em}

\textbf{Validity of Gumbel-Sinkhorn.} 
We test the advantage of applying the Gumbel-Sinkhorn technique to codewords selection. Three baselines to construct the sub-codebook are considered: (1) Directly selecting the top-$n$ frequent codewords from a learnt BNN (ReActNet-18). (2) Randomly selecting codewords. (3) Selecting codewords with an equal interval of indices, by setting $s_i=\lfloor\frac{i}{n-1}\rfloor\times (N-1)+1$ in Equation~\ref{eq:selection}.  Figure \ref{pic:ablation_study} (Left) shows that these baselines are much inferior to ours, indicating the superiority of our permutation learning. We observe a severe performance cut-down at 0.56-bit of the third baseline, reflecting the unpredictable performance variation for naive  selection methods.

\textbf{Comparison with product quantization.} Our proposed sub-codebook construction is related with but distinct from the product quantization~\cite{iclrStockJGGJ20,DBLP:journals/corr/abs-2010-02778} in two aspects: (1) As already specified in Figure~\ref{pic:distribution}, we construct the codebook with kernel-wise codewords rather than channel-wise ones. (2) We learn to jointly select codewords from a fixed codebook instead of directly optimizing the codewords independently, which  avoids the degeneration of the sub-codebook diversity. 
Figure~\ref{pic:ablation_study} (Middle) illustrates the comparisons regarding these two aspects. We observe  that  using kernel-wise codewords largely outperforms using the channel-wise codewords, and the selection-based optimization consistently achieves better performance than the product quantization (such as SNN~\cite{wang2021snn}) by a significant margin. In Figure~\ref{pic:ablation_study} (Right), the diversity of codewords is severely degenerated when applying the product quantization, whereas our method preserves the diversity.

\textbf{Convergence}. 
Figure~\ref{pic:patterns_step} displays the codewords selection process during training for 0.44-bit Sparks (ResNet-18) on ImageNet.
The selection initially fluctuates but converges in the end. This explains the convergence of our algorithm, which complies with our derivations in Theorem~\ref{thm:nonconvex}.

\textbf{Practical inference on FPGA}. We utilize the hardware framework SystemC-TLM\footnote{\url{https://www.accellera.org/community/systemc/about-systemc-tlm}} to model an FPGA. Our 0.56-bit Sparks achieves 1.167ms (ResNet-18) and 2.391ms (ResNet-34). The corresponding BNNs achieve 3.713ms and 7.806ms. Thus we have over three times acceleration.  

\vspace{-0.3em}
\section{Conclusion}
\vspace{-0.3em}
We propose a novel method named Sparse Kernel Selection (Sparks), which devises below 1-bit models by grouping kernels into a selected sub-codebook. The selection process is learnt end-to-end with Permutation Straight-Through Estimator (PSTE). Experiments show that our method is applicable for general tasks including image classification and object detection. One potential limitation of our research lies in that model compression requires the access of model parameters, leading to the  risk of unprotected  privacy.

\section*{Acknowledgement}
\small{
This work is funded by the Sino-German Collaborative Research Project Crossmodal Learning (NSFC 62061136001/DFG TRR169). W. Huang, F. Sun, and A. Yao are corresponding authors. Part of this work was done when the first author was an intern at Intel Labs China. Y. Wang and Y. Dong are supported by the Shuimu Tsinghua Scholar Program. }

\onecolumn

\clearpage 
\section*{\LARGE Appendix}
\appendix

\section{Proofs of Our Statements}
\label{sec:proof}
\setcounter{theorem}{0}
\setcounter{property}{0}
\begin{property}
We denote $\sB=\{-1,+1\}^{K\times K}$ as the dictionary of binary kernels. For each $\vw\in\R^{K\times K}$, the binary kernel $\hat{\vw}$ can be derived by a grouping process:
\begin{equation*}
\label{eq:grouping}
  \hat{\vw}= \mathrm{sign}(\vw)=\argmin_{\vu\in\sB}\|\vu-\vw\|_2.
\end{equation*}
\end{property}
\vskip-0.5em
\begin{proof}We denote $w(k_1,k_2)$ the entry of $\vw$ in the $k_1$-th row and $k_2$-th column, and similar denotation follows for $\vu$. We have,  
 \begin{align*}
  \mathop{\argmin}_{\vu\in\sB}\|\vu-\vw\|_2^2&=\mathop{\argmin}_{\vu\in\sB}\sum_{k_1,k_2}|u(k_1,k_2)-w(k_1,k_2)|^2\\
  &=\{\mathop{\argmin}_{u(k_1,k_2)\in \{-1,+1\} }|u(k_1,k_2)-w(k_1,k_2)|^2\}_{k_1,k_2}\\
  &=\{\mathrm{sign}(w(k_1,k_2))\}_{k_1,k_2}\\  
  &=\mathrm{sign}(\vw),
 \end{align*}
 which concludes the proof.
 \end{proof}

\begin{figure*}[h!]
\centering
\hskip-0.4em
\includegraphics[scale=0.32]{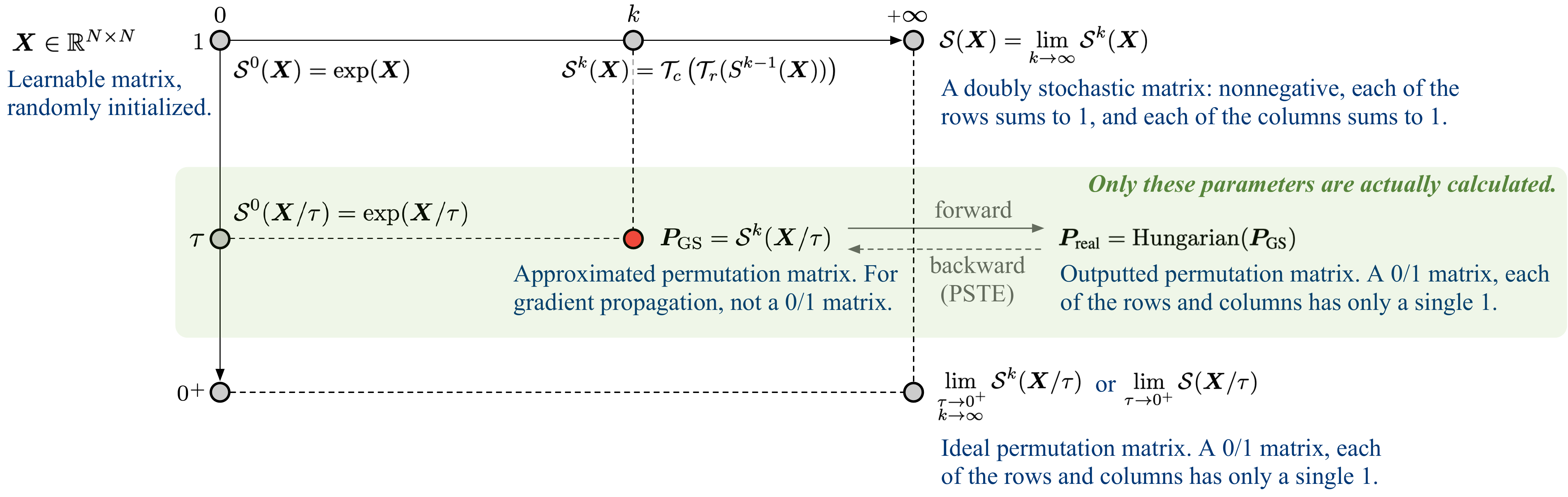}
\vskip-0.4em
\caption{Relationship of notations for a better understanding. The horizontal axis and vertical axis stand for the values of $k$ and $\tau$, respectively. The Gumbel noise $\epsilon$ is omitted here. Only parameters in the green region are actually calculated during training, while the others are only for the understanding purpose. All these parameters are NOT needed during inference. Starting from a learnable matrix $\bm{X}$, we adopt a temperature $\tau$ and calculate $\gS^0(\mX/\tau)$. We then perform the Sinkhorn operation for $k$ iterations to further obtain $\mP_{\mathrm{GS}}=\gS^k(\mX/\tau)$. We apply the Hungarian algorithm to further obtain $\Pr$. For sufficiently large $k$ and small $\tau$, $\Pr$ equals  the ideal permutation matrix. }
\label{pic:relation}
\end{figure*}

Before the proofs for the lemma  and theorem, in Figure~\ref{pic:relation}, we provide the relationship of notations that are adopted in the main paper to facilitate the understanding of our permutation learning process.

\setcounter{theorem}{0}
\begin{lemma}
\label{lemma1_proof}
For sufficiently large k and small $\tau$, we define the entropy of a doubly-stochastic matrix $\mP$ as $h(\mP)=-\sum_{i,j}P_{i,j}\log P_{i,j}$, and denote the rate of convergence for the Sinkhorn operator as $r\,(0<r<1)$\footnote{The Sinkhorn operator has a rate of convergence $r$ bounded by a value lower than 1 as proved by~\cite{journalssiammaxKnight08}.}. There exists a convergence series $s_\tau$ ($s_\tau\rightarrow0$ when $\tau\rightarrow0^+$) that satisfies  
\begin{align}
\|\Pr-\PGS\|_2^2 = \mathcal{O} \big(s_\tau^2+  r^{2k}\big).
\end{align}
\end{lemma}

\begin{proof}Let $\bm{X}_\tau=\bm{X}/\tau$ and $\bm{X}_0=\lim_{\tau\to0^+}\bm{X}_\tau$. In addition, follow the definitions in Equation~\ref{eq:sinkhorn}, we denote $\mathcal{S}^k(\cdot)$ the $k$-th iteration of the Sinkhorn output and $\mathcal{S}(\cdot)=\lim_{k\to \infty}\mathcal{S}^k(\cdot)$. 

As proved by~\cite{journalssiammaxKnight08}, the Sinkhorn operator has a rate of convergence $r\,(0<r<1)$ with respect to $k$, where $r$ always exists and is bounded by a value lower than 1. There is
\begin{align*}
\|\mathcal{S}^k(\bm{X}_\tau)-\mathcal{S}(\bm{X}_\tau)\|_2\leq r\|\mathcal{S}^{k-1}(\bm{X}_\tau)-\mathcal{S}(\bm{X}_\tau)\|_2\leq ,\cdots,\leq r^{k-1}\|\mathcal{S}^{1}(\bm{X}_\tau)-\mathcal{S}(\bm{X}_\tau)\|_2.
\end{align*}

By the definition of $\mathcal{S}^k(\bm{X}_\tau)$ in Equation~\ref{eq:sinkhorn}, the values of $\mathcal{S}^k(\bm{X}_\tau)$ are located in $[0,1]$ for all $k\geq1$. In addition, $\mathcal{S}(\bm{X}_\tau)$ is a $0/1$ matrix with exactly $N$ ones, where $N$ is the dimension. Therefore, $\|\mathcal{S}^{1}(\bm{X}_\tau)-\mathcal{S}(\bm{X}_\tau)\|_2^2$ is well bounded and  thus we obtain
\begin{align*}
\|\mathcal{S}^k(\bm{X}_\tau)-\mathcal{S}(\bm{X}_\tau)\|_2\leq C_1 r^{k},
\end{align*}
where $C_1>0$ is a constant.

As mentioned in Equation~\ref{eq:permutation}, $\mathcal{S}(\bm{X}_\tau)$ must be a doubly-stochastic matrix ~\cite{sinkhorn1964relationship}. According to Lemma 3 in~\cite{iclrMenaBLS18}, if denoting $f_0(\cdot)=\lin*{\cdot}{\bm{X}}_F$, there is $|f_0(\mathcal{S}(\bm{X}_0))-f_0(\mathcal{S}(\bm{X}_\tau))|\leq \tau (h(\mathcal{S}(\bm{X}_\tau))-h(\mathcal{S}(\bm{X}_0)))=\tau h(\mathcal{S}(\bm{X}_\tau))\leq\tau\max_{\mP\in\gB_N}(h(\mP))$, where $\gB_N$ denotes the the set of doubly stochastic matrices of dimension $N$. 

As proved by Lemma 3 in~\cite{iclrMenaBLS18}, $|f_0(\mathcal{S}(\bm{X}_0))-f_0(\mathcal{S}(\bm{X}_\tau))|\leq\tau\max_{\mP\in\gB_N}(h(\mP))$  implies the convergence of $\mathcal{S}(\bm{X}_\tau)$ to $\mathcal{S}(\bm{X}_0)$ and there exists a convergence series $s_\tau$ ($s_\tau\rightarrow0$ when $\tau\rightarrow0^+$), satisfying $\|\mathcal{S}(\bm{X}_0)-\mathcal{S}(\bm{X}_{\tau})\|_2\leq C_2 s_\tau$, where $C_2>0$ is a constant.

Based on the triangle inequality, there is
\begin{align*}
\|\mathcal{S}(\bm{X}_0)-\mathcal{S}^k(\bm{X}_\tau)\|_2^2\leq(C_2 s_\tau+ C_1 r^{k})^2 \leq 2C^2_2 s_\tau^2+ 2C^2_1 r^{2k}.
\end{align*}

As mentioned in \textsection~\ref{sec:pste}, there is $\PGS=\mathcal{S}^k(\bm{X}_\tau)$ if we omit the noise term. Given the convergence property, for sufficiently large $k$ and small $\tau$, the Hungarian algorithm output $\Pr$ equals to the real permutation, \emph{i.e.} $\Pr=\mathcal{S}(\bm{X}_0)$. In summary, we have
\begin{align*}
\|\Pr-\PGS\|_2^2 = \mathcal{O} \big(s_\tau^2+  r^{2k}\big),
\end{align*}
which concludes the proof.
\end{proof}

With the help of Lemma~\ref{lemma1_proof} and the inspiration of error-feedback framework~\cite{icmlKarimireddyRSJ19}, we now provide the detailed proof for Theorem~\ref{thm:nonconvex_proof}.

\setcounter{theorem1}{0}
\begin{theorem1}\label{thm:nonconvex_proof}  
Assume that the training objective $f$ \emph{w.r.t.}$\,\,\mP_{\mathrm{GS}}$ is $L$-smooth, and the stochastic gradient of $\mP_{\mathrm{real}}$ is bounded by $\mathbb{E}\|\mathbf{g}(\mP_{\mathrm{real}})\|_2^2\leq \sigma^2$. Denote the rate of convergence for the Sinkhorn operator as $r\,(0<r<1)$ and the stationary point as $\mP_{\mathrm{GS}}^{\star}$. Let the learning rate of PSTE be $\eta = \frac{c}{\sqrt{T}}$ with $c=\sqrt{\frac{f(\mP_{\mathrm{GS}}^0)-f(\mP_{\mathrm{GS}}^\star)}{{L \sigma^2}}}$. For a uniformly chosen $\bm{u}$ from the iterates $\{\Pr^0,\cdots,\Pr^T\}$, concretely $\bm{u}=\Pr^t$ with the probability $p_t=\frac{1}{T+1}$, it holds in expectation over the stochasticity and the selection of $\bm{u}$ :
\begin{small}
\begin{equation}
\mathbb{E}\|\nabla f(\bm{u})\|_2^2 = \mathcal{O} \left( \sigma\sqrt{\frac{f(\PGS^0)-f(\PGS^\star)}{T/L}}  + L^2\big(s_\tau^2 + r^{2k}\big) \right)\,.\vspace{-2mm}
\end{equation}\end{small}
\end{theorem1}

\begin{proof}Since the objective function $f$ is $L$-smooth, 
$\rvg({\PGS}) = \rvg({\Pr})$ in our PSTE, and $\PGS^{t+1} = \PGS^{t} - \eta\rvg(\Pr^{t})$, we can obtain the following derivations,
\begin{align*}
  f(\PGS^{t+1})&\leq f(\PGS^t) + \lin*{\PGS^{t+1} -\PGS^{t}}{\nabla f(\PGS^t)} + \frac{L}{2}\|\PGS^{t+1} -\PGS^{t}\|_2^2\\
  &= f(\PGS^t) - \eta\lin*{\mathbf{g}(\Pr^t)}{\nabla f(\PGS^t)} + \frac{L\eta^2}{2}\|\mathbf{g}(\Pr^t)\|_2^2\,.
\end{align*}

We use $\expect$ to represent the expectation with respect to the stochasticity. Based on the bound of the stochastic gradient, \emph{i.e.} $\mathbb{E}\|\mathbf{g}(\Pr^t)\|_2^2\leq \sigma^2$, and a natural property $\lin*{\mathbf{x}}{\mathbf{y}} \leq \frac{1}{2}\|\mathbf{x}\|_2^2+\frac{1}{2}\|\mathbf{y}\|_2^2$, it holds that,
\begin{align*}
  \expect\big[f(\PGS^{t+1}|\PGS^{t})\big]&\leq f(\PGS^t) - \eta\lin*{\mathbb{E}\big[\mathbf{g}(\Pr^t)\big]}{\nabla f(\PGS^t)} + \frac{L\eta^2}{2}\mathbb{E}\|\mathbf{g}(\Pr^t)\|_2^2\\
  &\leq f(\PGS^t) - \eta\lin*{\nabla f(\Pr^t)}{\nabla f(\PGS^t)} + \frac{L\eta^2\sigma^2}{2}\\
  &= f(\PGS^t) - \eta\lin*{\nabla f(\Pr^t)}{\nabla f(\Pr^t)} + \frac{L\eta^2\sigma^2}{2}+ \eta\lin*{\nabla f(\Pr^t)}{\nabla f(\Pr^t)-\nabla f(\PGS^t)} \\
  &\leq f(\PGS^t) - \eta\|\nabla f(\Pr^t)\|_2^2 + \frac{L\eta^2\sigma^2}{2}+\frac{\eta}{2}\|\nabla f(\Pr^t)\|_2^2+ \frac{\eta}{2}\|\nabla f(\Pr^t)-\nabla f(\PGS^t)\|_2^2\\
  &\leq f(\PGS^t)- \frac{\eta}{2}\|\nabla f(\Pr^t)\|_2^2 + \frac{L\eta^2\sigma^2}{2}+ \frac{\eta L^2}{2}\|\Pr^t-\PGS^t\|_2^2
  \,.
\end{align*}

With Lemma~\ref{lemma1_proof}, 
\begin{align*}
\expect\big[f(\PGS^{t+1}|\PGS^{t})\big] \leq f(\PGS^t) - \frac{\eta}{2}\|\nabla f(\Pr^t)\|_2^2 + \frac{L\eta^2\sigma^2}{2}+ \frac{\eta L^2}{2}\big(C_1 s_\tau^2 + C_2r^{2k}\big)\,,
\end{align*}
where $C_1>0$ and $C_2>0$ are constants.

By rearranging the orders and further applying the expectation on $\PGS^{t}$, 
\begin{align*}
\mathbb{E}\|\nabla f(\Pr^t)\|_2^2 \leq \frac{2}{\eta}\big(\expect\big[f(\PGS^{t})\big]-\expect\big[f(\PGS^{t+1})\big]  \big)+ L\eta\sigma^2+L^2\big(C_1 s_\tau^2 + C_2r^{2k}\big) \,.\vspace{-2mm}
\end{align*}

Summing over $t=0,1,\cdots,T$, 
\begin{align*}
\sum_{t=0}^T\mathbb{E}\|\nabla f(\Pr^t)\|_2^2 &\leq \frac{2}{\eta}\sum_{t=0}^T\big(\expect\big[f(\PGS^{t})\big]-\expect\big[f(\PGS^{t+1})\big]  \big)+ L\eta\sigma^2+L^2\sum_{t=0}^T\big(C_1 s_\tau^2 + C_2r^{2k}\big) \\
  &\leq \frac{2}{\eta}\big(f(\PGS^{0})-f(\PGS^{\star})  \big)+ (T+1)L\eta\sigma^2+L^2\sum_{t=0}^T\big(C_1 s_\tau^2 + C_2r^{2k}\big) \,.\vspace{-2mm}
\end{align*}

For a uniformly chosen $\bm{u}$ from the iterates $\{\Pr^0,\cdots,\Pr^T\}$, concretely $\bm{u}=\Pr^t$ with the probability $p_t=\frac{1}{T+1}$. Divide the inequation by $T+1$, and extend $\mathbb{E}$ to represent the expectation over the stochasticity and the selection of $\bm{u}$, there is
\begin{align*}
\mathbb{E}\|\nabla f(\bm{u})\|_2^2 \leq \frac{2}{\eta(T+1)}\big(f(\PGS^{0})-f(\PGS^{\star})  \big)+ L\eta\sigma^2+L^2\big(C_1 s_\tau^2 + C_2r^{2k}\big) \,.
\vspace{-2mm}
\end{align*}

Substituting the learning rate $\eta$, we finally obtain
\begin{align*}
\mathbb{E}\|\nabla f(\bm{u})\|_2^2 & \leq \frac{2\sigma\sqrt {LT}}{T+1}\sqrt{f(\PGS^0)-f(\PGS^\star)} + \frac{\sigma\sqrt L}{\sqrt {T}}\sqrt{f(\PGS^0)-f(\PGS^\star)}+L^2\big(C_1 s_\tau^2 + C_2r^{2k}\big) \,.\vspace{-2mm}\\
   &\leq 3\sigma\sqrt{\frac{f(\PGS^0)-f(\PGS^\star)}{T/L}}+L^2\big(C_1 s_\tau^2 + C_2r^{2k}\big) \,,
\end{align*}

Therefore,
\begin{align}
&\nonumber 
\mathbb{E}\|\nabla f(\bm{u})\|_2^2 = \mathcal{O} \left( \sigma\sqrt{\frac{f(\PGS^0)-f(\PGS^\star)}{T/L}}  + L^2\big(s_\tau^2 + r^{2k}\big) \right)\,,
\end{align}
which concludes the proof.
\end{proof}

\begin{figure*}[b]
\centering
\vskip-0.2em
\hskip-0.2em
\includegraphics[scale=0.4]{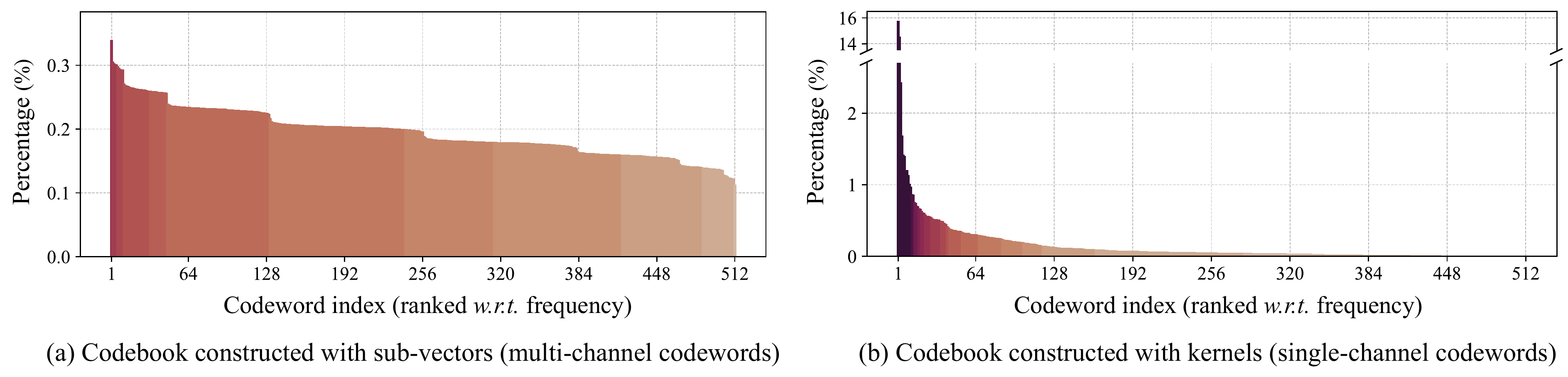}
\caption{A supplement to Figure~\ref{pic:distribution} by ranking codewords \emph{w.r.t.} the frequency. It shows that the ranked codewords in (b) nearly follow the power-law distribution.}
\label{pic:distribution_power}
\end{figure*}

\section{Experiment Details and Analysis}
\subsection{Implementation Details}

\textbf{Implementation of ImageNet training.} We follow the two-step scheme (as detailed in \textsection~\ref{sec:exp}) and the training settings in~\cite{DBLP:conf/eccv/LiuSSC20}. Specifically, for each step, the model is trained for 640k training iterations with batch size 512. We adopt the Adam optimizer~\cite{DBLP:journals/corr/KingmaB14} and set the initial learning rate to $10^{-3}$. Weight decay rates in the first and second steps are  $10^{-5}$ and $0$, respectively. For experiments on ImageNet, models are trained with 8 V100 GPUs. We follow the training settings and data augmentation strategies in~\cite{DBLP:conf/eccv/LiuSSC20}. 

\textbf{Implementation of CIFAR10 training.} 
For experiments on CIFAR10, each experiment is performed on a single V100 GPU. We train the network with 256 epochs in each step. We set the batch size to 256  and use an Adam optimizer~\cite{DBLP:journals/corr/KingmaB14}. The learning rate is initialized to $5\times10^{-4}$ and is updated by a linear learning rate decay scheduler. Results of our method on CIFAR10 are averaged over three runs.

\textbf{Implementation of  selection and ablation studies.} 
\label{sec:appendix_imple}
We  further detail our implementation of the codeword selection process based on Figure~\ref{pic:ablation_study} (Middle), where we provide four experiment settings including using kernel-wise/channel-wise codewords, and selection-based/product-quantization-based learning. Implementation details for these four experiments (labeled as (a)-(d), respectively) are described as follows. 
\begin{itemize}
\item[(a)] \emph{Selection, kernel-wise (our proposed method):} each codeword is a $3\times 3$ convolutional kernel with values being $\pm1$. We constantly keep the 1$^{\text{st}}$ (all $-1$s) and 512$^{\text{th}}$ (all $+1$s) codewords in the sub-codebook, as these two codewords take a large proportion. We divide the remaining 510 codewords into two halves (1$^{\text{st}}$ half: from index 2 to 256; 2$^{\text{nd}}$ half: from index 257 to 511). Obviously, each codeword in one half has a corresponding  codeword with opposite signs in another half. This technique speeds up the selection process without noticeably affecting the performance. 

\item[(b)]  \emph{Selection, channel-wise:} each codeword is a $1\times 9$ sub-vector with values being $\pm1$, and codewords are obtained from  flattened convolutional weights. We follow the speed up strategy in (a) by dividing codewords into two parts. Unlike (a), we do not constantly send the 1$^{\text{st}}$ (all $-1$s) and 512$^{\text{th}}$ (all $+1$s) codewords to the sub-codebook, since this process does not bring improvements for the channel-wise setting.

\item[(c)]  \emph{Product quantization, kernel-wise:} we follow the common method of product quantization~\cite{iclrStockJGGJ20,DBLP:journals/corr/abs-2010-02778}  to learn real-valued codewords. Before training, we randomly initialize $2^n$ different $3\times3$ real-valued codewords, with their values being $\pm1.0$. During training, in the forward pass, we obtain the binary codewords by applying  the $\mathrm{sign}$ function on the real-valued codewords. In the backward pass, the Straight-Through Estimator technique is adopted which copies the gradients of binary codebooks to real-valued codebooks. 

\item[(d)] \emph{Product quantization, channel-wise:} the learning process follows (c), yet each codeword is a $1\times 9$ sub-vector across multiple channels, obtained from  flattened convolutional weights.
\end{itemize}

\begin{table*}[b!]
\vskip -0.05in
\caption{Calculation details for the storage and BOPs as reported in Table~\ref{table:imagenet}. Networks are evaluated on   ImageNet  with ResNet-18.}
\vskip -0.03in
\label{table:calculation_details}
\vspace{-0.1in}
\begin{center}
\begin{small}
\tablestyle{2pt}{1.05}
\resizebox{0.99\linewidth}{!}{
\begin{tabular}{lcccccccc|cccc|cccc}
\toprule
\multirow{2}*{layer-name} & \multirow{2}*{input-w} & \multirow{2}*{input-h} & \multirow{2}*{input-c} & \multirow{2}*{output-w} & \multirow{2}*{output-h} & \multirow{2}*{output-c} & \multirow{2}*{kernel-w}& \multirow{2}*{kernel-h} & \multicolumn{4}{c|}{Storage (bit)} &\multicolumn{4}{c}{BOPs}\\ 
 &  &  &  &  &  &  &  & &1-bit (Base) & 0.78-bit & 0.67-bit & 0.56-bit & 1-bit (Base) & 0.78-bit & 0.67-bit & 0.56-bit\\
\cmidrule(r){1-17}
\multicolumn{1}{c}{\textcolor{Gray}A}&\textcolor{Gray}B&\textcolor{Gray}C&\textcolor{Cerulean}D&\textcolor{BurntOrange}E&\textcolor{BurntOrange}F&\textcolor{LimeGreen}G&\textcolor{Red}H&\textcolor{Red}I&\textcolor{ForestGreen}J&\textcolor{Purple}K&\textcolor{Purple}L&\textcolor{Purple}M&\textcolor{Salmon}N&\textcolor{SkyBlue}O&\textcolor{SkyBlue}P&\textcolor{SkyBlue}Q\\
\cmidrule(r){1-17}
conv1 & 224 & 224 & 3 & 112 & 112 & 64 & 7&7 & - & - & - & - & - & - & - & -\\
conv2-1a & 56 & 56 & 64 & 56 & 56 & 64 & 3&3 & 36864 & 28672 & 24576 & 20480 & 115605504 & 115605504 & 115605504 & 64225248\\
conv2-1b & 56 & 56 & 64 & 56 & 56 & 64 & 3&3 & 36864 & 28672 & 24576 & 20480 & 115605504 & 115605504 & 115605504 & 64225248\\
conv2-2a & 56 & 56 & 64 & 56 & 56 & 64 & 3&3 & 36864 & 28672 & 24576 & 20480 & 115605504 & 115605504 & 115605504 & 64225248\\
conv2-2b & 56 & 56 & 64 & 56 & 56 & 64 & 3&3 & 36864 & 28672 & 24576 & 20480 & 115605504 & 115605504 & 115605504 & 64225248\\
conv3-1a & 56 & 56 & 64 & 28 & 28 & 128 & 3&3 & 73728 & 57344 & 49152 & 40960 & 57802752 & 57802752 & 32112576 & 17661888\\
conv3-1b & 28 & 28 & 128 & 28 & 28 & 128 & 3&3 & 147456 & 114688 & 98304 & 81920 & 115605504 & 115605504 & 64225216 & 35323840\\
conv3-2a & 28 & 28 & 128 & 28 & 28 & 128 & 3&3 & 147456 & 114688 & 98304 & 81920 & 115605504 & 115605504 & 64225216 & 35323840\\
conv3-2b & 28 & 28 & 128 & 28 & 28 & 128 & 3&3 & 147456 & 114688 & 98304 & 81920 & 115605504 & 115605504 & 64225216 & 35323840\\
conv4-1a & 28 & 28 & 128 & 14 & 14 & 256 & 3&3 & 294912 & 229376 & 196608 & 163840 & 57802752 & 32112512 & 17661824 & 10436480\\
conv4-1b & 14 & 14 & 256 & 14 & 14 & 256 & 3&3 & 589824 & 458752 & 393216 & 327680 & 115605504 & 64225152 & 35323776 & 20873088\\
conv4-2a & 14 & 14 & 256 & 14 & 14 & 256 & 3&3 & 589824 & 458752 & 393216 & 327680 & 115605504 & 64225152 & 35323776 & 20873088\\
conv4-2b & 14 & 14 & 256 & 14 & 14 & 256 & 3&3 & 589824 & 458752 & 393216 & 327680 & 115605504 & 64225152 & 35323776 & 20873088\\
conv5-1a & 14 & 14 & 256 & 7 & 7 & 512 & 3&3 & 1179648 & 917504 & 786432 & 655360 & 57802752 & 17661696 & 10436352 & 6823680\\
conv5-1b & 7 & 7 & 512 & 7 & 7 & 512 & 3&3 & 2359296 & 1835008 & 1572864 & 1310720 & 115605504 & 35323648 & 20872960 & 13647616\\
conv5-2a & 7 & 7 & 512 & 7 & 7 & 512 & 3&3 & 2359296 & 1835008 & 1572864 & 1310720 & 115605504 & 35323648 & 20872960 & 13647616\\
conv5-2b & 7 & 7 & 512 & 7 & 7 & 512 & 3&3 & 2359296 & 1835008 & 1572864 & 1310720 & 115605504 & 35323648 & 20872960 & 13647616\\
fc1000 & 1 & 1 & 512 & 1 & 1 & 1000 & -&- & - & - & - & - & - & - & - & -\\
\cmidrule(r){1-17}
\multicolumn{9}{c|}{\textbf{Total}}& \makecell{10985472\\=11.0Mbit} & \makecell{8544256\\=8.57Mbit} & \makecell{7323648\\=7.32Mbit} & \makecell{6103040\\=6.10Mbit} & \makecell{1676279808\\=1.68$\times10^9$} & \makecell{1215461888\\=1.22$\times10^9$} & \makecell{883898624\\=0.88$\times10^9$} & \makecell{501356672\\=0.50$\times10^9$}\\
\bottomrule
\end{tabular}}
\end{small}
\end{center}
\begin{tablenotes}
\vspace{-0.02in}
    \footnotesize
    \item[1] \scriptsize{\textbf{Storage:} \textcolor{ForestGreen}J$=$\textcolor{LimeGreen}G$\times$\textcolor{Cerulean}D$\times$\textcolor{Red}H$\times$\textcolor{Red}I$;$ \textcolor{Purple}K$=$\textcolor{ForestGreen}J$\times\log_2(128)/9;$ \textcolor{Purple}L$=$\textcolor{ForestGreen}J$\times\log_2(64)/9;$ \textcolor{Purple}M$=$\textcolor{ForestGreen}J$\times\log_2(32)/9$.}
    \vskip0.03in
    \item[2] \scriptsize{\textbf{BOPs:} 
    \textcolor{Salmon}N$=$\textcolor{Cerulean}D$\times$\textcolor{BurntOrange}E$\times$\textcolor{BurntOrange}F$\times$\textcolor{Red}H$\times$\textcolor{Red}I$\times$\textcolor{LimeGreen}G$;$ 
    \textcolor{SkyBlue}O$=\hskip-0.3em\min\{$\textcolor{Salmon}N$,$ \textcolor{Salmon}N$/$\textcolor{LimeGreen}G$\times128+$\textcolor{LimeGreen}G$\times ($\textcolor{Cerulean}D$\times$\textcolor{BurntOrange}E$\times$\textcolor{BurntOrange}F$ -1)/2\}$; 
    \textcolor{SkyBlue}P$=\hskip-0.3em\min\{$\textcolor{Salmon}N$,$ \textcolor{Salmon}N$/$\textcolor{LimeGreen}G$\times64+$\textcolor{LimeGreen}G$\times($\textcolor{Cerulean}D$\times$\textcolor{BurntOrange}E$\times$\textcolor{BurntOrange}F$ -1)/2\}$; 
    \textcolor{SkyBlue}Q$=\hskip-0.3em\min\{$\textcolor{Salmon}N$,$ \textcolor{Salmon}N$/$\textcolor{LimeGreen}G$\times32+$  \textcolor{LimeGreen}G$\times($\textcolor{Cerulean}D$\times$\textcolor{BurntOrange}E$\times$\textcolor{BurntOrange}F$ -1)/2\}$.}
\end{tablenotes}
\end{table*}

\textbf{Storage and BOPs calculation.} In Table~\ref{table:calculation_details}, we provide  details of how storages and BOPs in Table~\ref{table:imagenet} are calculated. The calculation follows the analysis  described in \textsection~\ref{sec:complexity}.

\begin{figure*}[t]
\centering
\hskip-0.3em
\includegraphics[scale=0.42]{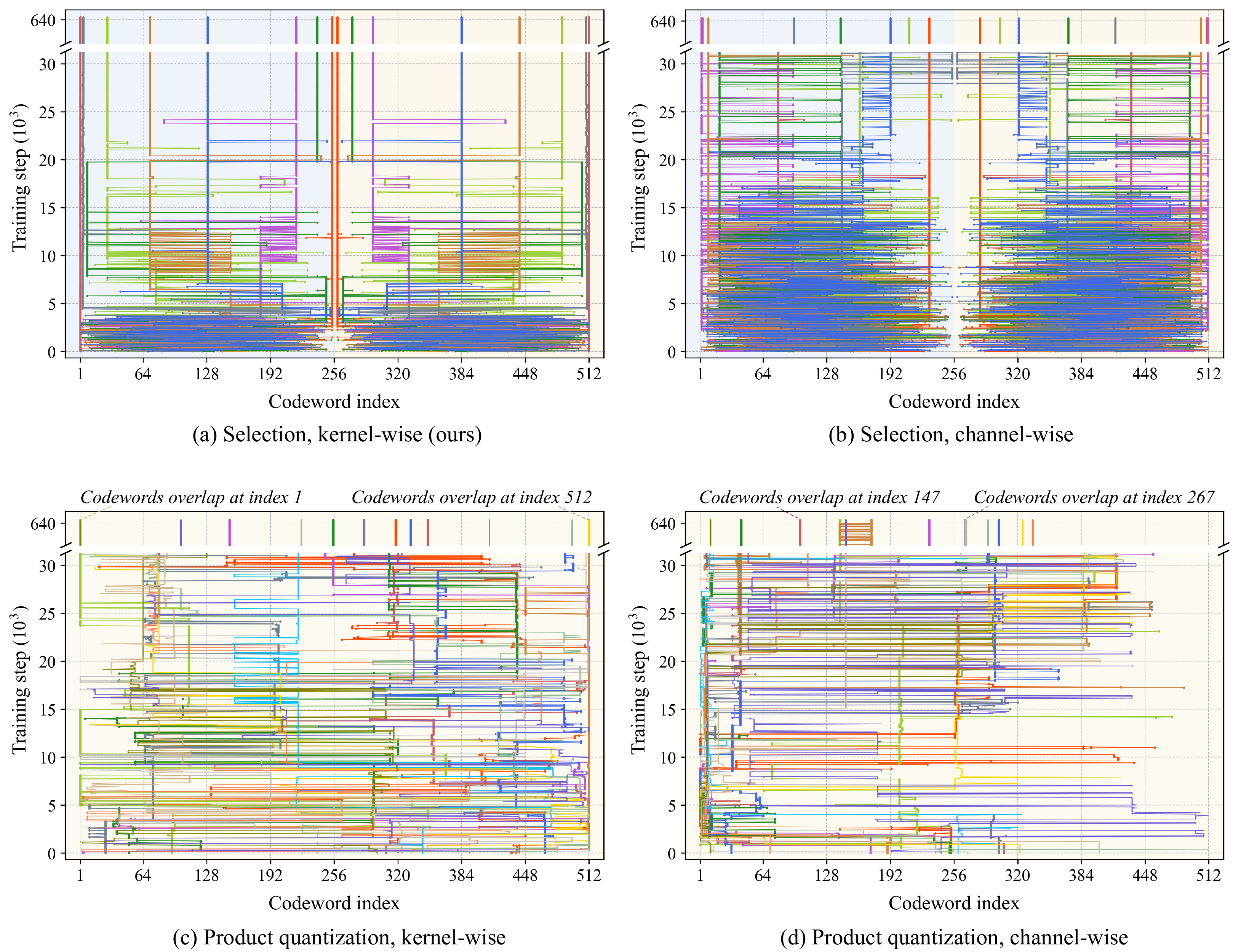}
\caption{Comparison of codewords learning processes when using kernel-wise/channel-wise codewords and selection-based/product-quantization-based learning. The corresponding experimental results are already provided in Figure~\ref{pic:ablation_study} (Middle). All experiments are based on 0.44-bit settings with $n=16$, and are performed on ImageNet upon ResNet-18. We observe that, when both using selection-based learning, kernel-wise codewords in (a) converge much faster (within $25\times10^3$ steps) than channel-wise codewords in (b); codewords with product-quantization-based learning in (c) and (d) also converge slower than (a), and are likely to overlap during training which degenerates the codebook diversity.}
\label{pic:pattern_steps_four_figs}
\vskip0.7em
\end{figure*}

\subsection{Additional Experiment Analysis}

\textbf{Power-law property.} Figure~\ref{pic:distribution_power} illustrates the distributions when ranking codewords in Figure~\ref{pic:distribution} according to the codeword frequency. It shows that in Figure~\ref{pic:distribution_power}(b), codewords nearly obey the power-law distribution.

\begin{figure*}[t]
\centering
\vskip-0.6em
\includegraphics[scale=0.425]{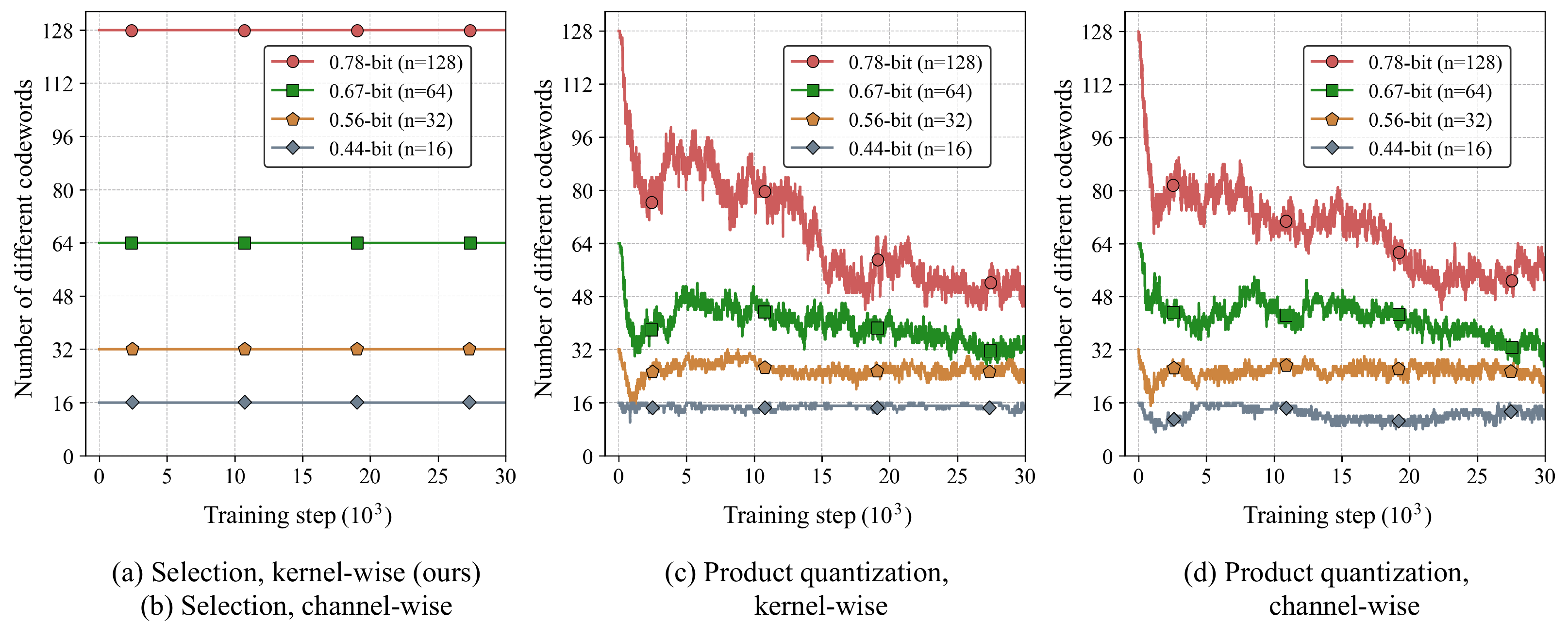}
\vskip-0.05em
\caption{Numbers of different codewords when using  selection-based/product-quantization-based learning, as a supplement to Figure~\ref{pic:ablation_study} (Right). (a)-(d) correspond to the experiments in Figure~\ref{pic:ablation_study}  (Right) and Figure~\ref{pic:pattern_steps_four_figs}. Experiments are performed on ImageNet upon ResNet-18. We provide four settings, including 0.78-bit, 0.67-bit, 0.56-bit and 0.44-bit, with $n=\,$128, 64, 32 and 16, respectively. We observe that the sub-codebook highly degenerates during training in (c) and (d), since codewords tend to be repetitive when being updated independently. While in (a) and (b), the diversity of codewords preserves, which implies the superiority of our selection-based learning.}
\label{pic:num_decay}
\end{figure*}

\textbf{Codewords selection and overlaps.} In Figure~\ref{pic:pattern_steps_four_figs}, we further compare the   codewords learning processes for the four settings in Figure~\ref{pic:ablation_study} (Middle). As a supplement to Figure~\ref{pic:ablation_study} (Right), Figure~\ref{pic:num_decay} provides the change of sub-codebooks during  training of the four experiment settings (a)-(d). As codewords in (c) and (d) tend to overlap during training, the diversity is severely affected. In addition, we also conduct experiments that at each step or every several steps, we replenish the sub-codebook with random different codewords so that the sub-codebook size recovers to $2^n$, but the performance is very close to directly selecting codewords at random (as already shown in Figure~\ref{pic:ablation_study} (Left), randomly selection achieves low performance).

\textbf{Acceleration of training.} We consider two approaches that can accelerate the training process on large datasets (\emph{e.g.} ImageNet), without causing much detriment to the performance. (1) We conduct permutation learning only in the first $30\times10^3$ training steps, and fix the selected codewords for the later training. (2) We obtain the sub-codebook by pretraining on a small dataset like CIFAR10, and directly adopt the sub-codebook for ImageNet without further permutation learning. Compared with the results of our Sparks reported in Table~\ref{table:imagenet}, the performance does not decrease when using the acceleration approach (1), and decreases slightly ($-0.4$\%, $-0.6$\%, and $-1.1$\% for 0.78-bit, 0.67-bit, and 0.56-bit, respectively) when using the approach (2).

\begin{figure*}[t]
\centering
\includegraphics[scale=0.65]{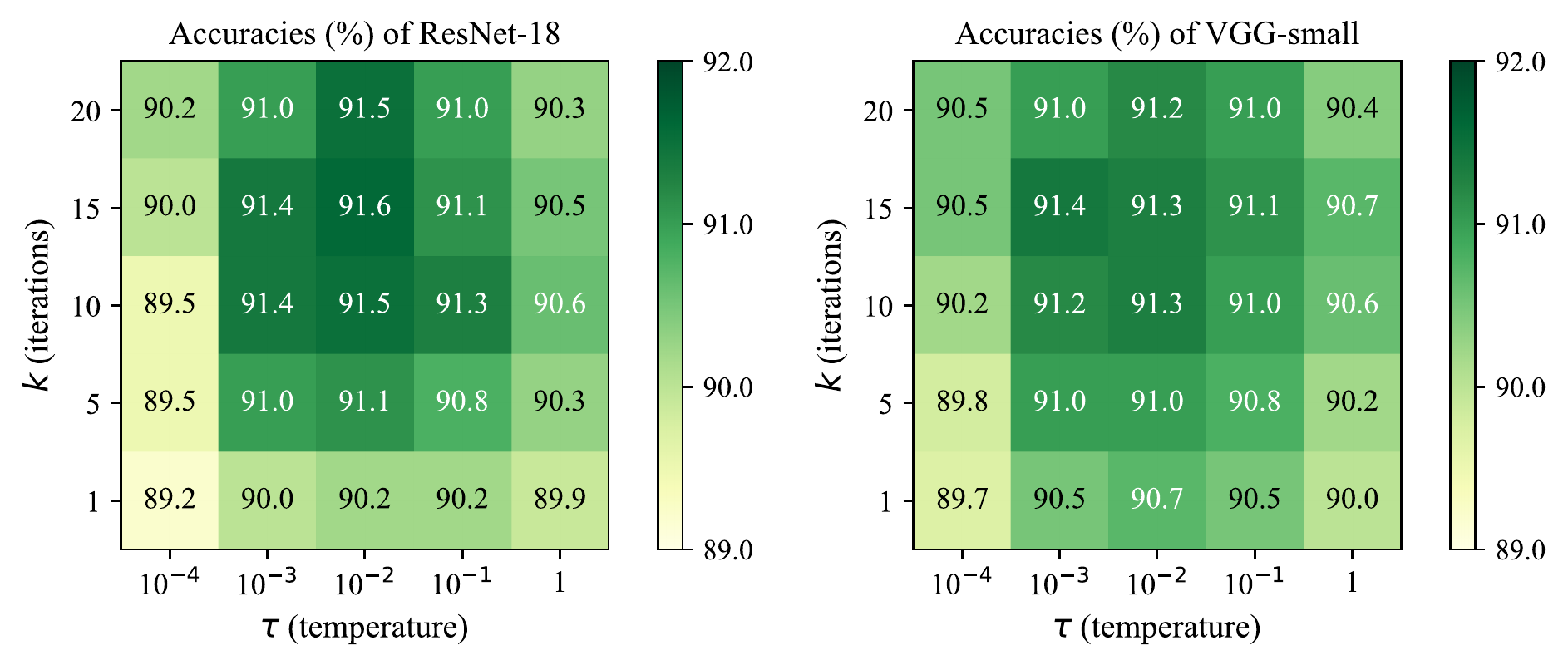}
\vskip-0.25em
\caption{Sensitivity analysis for hyper-parameters including the iteration number $k$ and the temperature $\tau$, as adopted in Equation~\ref{eq:selection-final} and also clearly illustrated in Figure~\ref{pic:relation}. Experiments are conducted on the CIFAR10 dataset with 0.56-bit ResNet-18 and VGG-small, respectively. Results are averaged over three runs with different random seeds. Both hyper-parameters are insensitive around the default values $k=10$ and $\tau=10^{-2}$.}
\label{pic:sensitivity}
\vskip-0.1em
\end{figure*}

\textbf{Sensitivity analysis of hyper-parameters.} 
In Figure \ref{pic:sensitivity}, we compare the accuracy with different hyper-parameter settings for $k$ and $\tau$  in Equation~\ref{eq:selection-final}. In the PSTE optimization, $k$ is the iteration number and $\tau$ is the temperature. Experiments are performed with ResNet-18 and VGG-small on CIFAR10. We observe that both hyper-parameters are insensitive around their default settings $k=10$ and $\tau=10^{-2}$. The parameter $k$ is quite easy to choose since results are stable when $k=5\sim20$. In addition, regarding two extreme cases, 
setting $\tau$ too small (\emph{e.g.} $10^{-4}$) will hinder the smoothness of gradient back-propagation, and assigning a too large value (\emph{e.g.} $1$) will enlarge the permutation approximation error, both of which may cause detriment to the final performance. Luckily, according to Figure \ref{pic:sensitivity}, the performance is stable when changing $\tau$ by $10$ or $1/10$ times around the default value, implying the high stability of our method.

\textbf{About top-n most frequent codewords:} In Figure~\ref{pic:ablation_study}, we compare sampling top-n most frequent codewords and our method. We display an example of $0.44$-bit top-n codewords here: $\vcenter{\hbox{\includegraphics[width=17em, height=2.3ex]{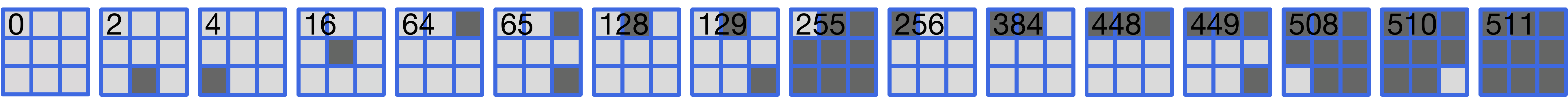}}}$.  It shows the top-n codewords tend to choose adjacent codewords more frequently, which could hinder the diversity of the codebook. By contrast, as shown in Figure~\ref{pic:patterns_step}, our learned $0.44$-bit method outputs diverse codewords, yielding better performance particularly at $0.56$-bit from 61.7 to 64.3. 

\textbf{A two-step recipe with product quantization.}  Given the purpose of attaining a compact BNN, we also conduct an intuitive two-step baseline method: at first, load parameters of a standard BNN (ReActNet-18, from the open-sourced model) as the pre-trained model; then perform product quantization on the binary weights to compress the BNN. By this method, we achieve 59.1\% accuracy on ImageNet under 0.56-bit, and the sub-codebook degeneration still exists. Such result is much inferior to our Sparks under 0.56-bit (64.3\%).

\begin{table}[b!]
	\centering
\vskip -0.07in
	\tablestyle{1pt}{0.8}
	\caption{{The definition and usage of  important symbols.}}

	\resizebox{0.9\linewidth}{!}{
		\begin{tabular}{ll}
			\toprule
			Symbol & Definition and Usage\\
			\midrule
			$\vw\in\R^{K\times K}$&Convolutional kernel with the kernel size $K$.  \\
			$\hat{\vw}\in\{-1,+1\}^{K\times K}$&Selected binary kernel for $\vw$. There is $\hat{\vw}\in\text{sub-codebook }\sU\subseteq\text{codebook }\sB=\{-1,+1\}^{K\times K}$.\\
			$N\in\mathbb{Z}^+$&$N=|\sB|$, the codebook size, $N=512$ when $K=3$.\\
			$n\in\mathbb{Z}^+$&$n=|\sU|$, the sub-codebook size, \emph{e.g.}, for $0.56$-bit Sparks, $n=32$.\\
			$\mB\in\{\pm1\}^{K^2\times N}$&  Column by column indexing from $|\sB|$.\\
			$\mU\in\{\pm1\}^{K^2\times n}$& Column by column indexing from $|\sU|$.\\
			$\mV\in\{0,1\}^{N\times n}$& A pre-defined selection matrix.\\
			$k\in\mathbb{Z}^+$& The number of iteration to approximate the permutation matrix in Equation~\ref{eq:selection-final}. \\
			$\tau\in\R^+$& A small temperature to approximate the permutation matrix in Equation~\ref{eq:selection-final}.\\
			$\mX\in\R^{N\times N} $& A randomly initialized, learnable matrix.\\
			$\mP_{\mathrm{GS}}\in\R^{N\times N}$& $\mP_{\mathrm{GS}}=\gS^k((\mX+\epsilon)/\tau)$, the approximated permutation matrix for  propagation, not a $0/1$ matrix.\\
			$\mP_{\mathrm{real}}\in\{0,1\}^{N\times N}$&$\mP_{\mathrm{real}}= \mathrm{Hungarian}(\mP_{\mathrm{GS}})$, the outputted permutation matrix, a doubly stochastic $0/1$ matrix.\\
			\bottomrule
		\end{tabular}
	}
\vskip -0.07in
\label{table:symbol}
\end{table}

\textbf{Symbol table.} We list  the definition and usage  important symbols in Table~\ref{table:symbol}.

\textbf{Whether  indexing process hinders valid acceleration:} No, the  indexing process of binary kernels does not hinder valid acceleration. {{(1)}} Indexing $n$ codewords is very cheap, \emph{e.g.}, only $32$ codewords for our $0.56$-bit model ($<0.5$ nanosecond on our FPGA). {{(2)}} Indexing $3\times 3$ pre-calculated results ($3\times 3$ codeword $\circledast$ $3\times 3$ feature region, see \textsection~\ref{sec:complexity}) is also negligible based on our implementation with a Lookup Table (LUT) that stores $3\times 3$ pre-calculated results: The practical LUT size, instead of $n \times C_{\text{in}}\times H\times W$ by dividing input feature maps into {$1\times 3\times 3$ slices}, and sending only one slice  to a Processing Engine (PE) at each clock cycle. This leads to very low latency for the lookup process, \emph{e.g.}, LUT size is $32$ for our $0.56$-bit model ($<0.5$ nanosecond for indexing), which is easily implemented in the current clock cycle.

\section{Object Detection}
\vspace{-0.2em}
\subsection{Implementation}
We evaluate our method for object detection on two benchmark datasets: PASCAL VOC~\cite{ijcvEveringhamGWWZ10} and COCO~\cite{eccvLinMBHPRDZ14}. We follow the standard data split settings~\cite{cvprWangWL020}. Regarding the PASCAL VOC dataset, we train our model on both the VOC 2007 and VOC 2012 trainval sets, which together contain about 16k natural images of 20 different categories in total. We evaluate the performance on VOC 2007 test set that is composed of about 5k images. 
COCO dataset (2014 object detection track) is a large-scale dataset that collects images from 80 different categories. We train our model with 80k training images as well as 35k images sampled from the validation set (denoted as trainval35k~\cite{cvprBellZBG16}), and carry out evaluations on the remaining 5k images in the validation set (minival~\cite{cvprBellZBG16}). 

We follow BiDet~\cite{cvprWangWL020} for the basic training settings including parameters and data augmentation methods. Specifically, we train 50 epochs in total with  batch size 32 and the Adam optimizer. We initialize the learning rate to $10^{-3}$ which decays by multiplying 0.1 at the 6$^{\text{th}}$ and 10$^{\text{th}}$ epoch.  We consider two typical architectures including SSD300~\cite{DBLP:conf/eccv/LiuAESRFB16} (with VGG-16~\cite{DBLP:journals/corr/SimonyanZ14a}) and Faster R-CNN~\cite{DBLP:conf/nips/RenHGS15} (with ResNet-18~\cite{he2016deep}) to verify the effectiveness and generalization of our method.

\begin{table*}[t!]
\caption{Performance comparisons with object detection methods on the PASCAL VOC dataset. Sparks$^*$ indicates using the two-step training method and the generalized Sign/PReLU functions (as adopted in \cite{DBLP:conf/eccv/LiuSSC20} for image classification).}
\vskip-0.1in
\label{table:object_voc}
\begin{center}
\begin{small}
\tablestyle{4.5pt}{1.0}
\resizebox{0.9\linewidth}{!}{
	\begin{tabular}{lcccc|lcccc}
		\toprule
		{Method}&Bit-width& mAP &Storage&BOPs&{Method}&Bit-width& mAP  &Storage&BOPs\\
		\bf{(SSD300)}&(W$/$A) & (\%)&Saving&Saving&\bf{(Faster R-CNN)}&(W$/$A) &(\%)& Saving&Saving\\
		\midrule
		Full-precision &32$/$32&72.4 & 1$\times$&1$\times$ & Full-precision &32$/$32&74.5 & 1$\times$&1$\times$\\
		\cmidrule(r){1-10}
		BNN~\cite{nipsHubaraCSEB16}&1$/$1&42.0& 32$\times$&64$\times$&BNN~\cite{nipsHubaraCSEB16}&1$/$1&35.6& 32$\times$&64$\times$\\
		XNOR-Net~\cite{eccvRastegariORF16}&1$/$1&50.2& 32$\times$&64$\times$&XNOR-Net~\cite{eccvRastegariORF16}&1$/$1&48.4& 32$\times$&64$\times$\\
		Bi-RealNet~\cite{ijcvLiuLWYLC20}&1$/$1&63.8& 32$\times$&64$\times$&Bi-RealNet~\cite{ijcvLiuLWYLC20}&1$/$1&58.2& 32$\times$&64$\times$\\
		BiDet~\cite{cvprWangWL020}&1$/$1&66.0& 32$\times$&64$\times$&BiDet~\cite{cvprWangWL020}&1$/$1&59.5& 32$\times$&64$\times$\\
		\cmidrule(r){1-10}
		Sparks (ours) & 0.78$/$1 &65.2  &  41.0$\times$&108$\times$&Sparks (ours) & 0.78$/$1 &58.9  &  41.0$\times$&88$\times$\\
		Sparks (ours) & 0.56$/$1 & 64.3 & 57.1$\times$&285$\times$&Sparks (ours) & 0.56$/$1 & 57.7 & 57.1$\times$&214$\times$\\
		\cmidrule(r){1-10}
		Sparks$^*$ (ours) & 0.78$/$1 &68.9  &  41.0$\times$&108$\times$&Sparks$^*$ (ours) & 0.78$/$1 &66.2  &  41.0$\times$&88$\times$\\
		Sparks$^*$ (ours) & 0.56$/$1 & 68.0 & 57.1$\times$&285$\times$&Sparks$^*$ (ours) & 0.56$/$1 & 65.5 & 57.1$\times$&214$\times$\\
		\bottomrule
	\end{tabular}}
\end{small}
\end{center}
\vskip-0.5em
\end{table*}

\begin{table*}[t!]
\caption{Performance comparisons with object detection methods on the COCO dataset. Sparks$^*$ indicates using the two-step training method and the generalized  Sign/PReLU functions (as adopted in \cite{DBLP:conf/eccv/LiuSSC20} for image classification).}
\vskip-0.1in
\label{table:object_coco}
\begin{center}
\begin{small}
\tablestyle{1.5pt}{1.19}
\resizebox{0.9\linewidth}{!}{
	\begin{tabular}{lcccccccp{0.005cm}|p{0.005cm}lccccccc}
		\toprule
		{Method}&Bit-width& mAP & AP$_{50}$& AP$_{75}$& AP$_{s}$& AP$_{m}$& AP$_{l}$&&&{Method}&Bit-width&mAP & AP$_{50}$& AP$_{75}$& AP$_{s}$& AP$_{m}$& AP$_{l}$\\
		\bf{(SSD300)}&(W$/$A) & (\%)& (\%)& (\%)& (\%)& (\%)& (\%)&&& \bf{(Faster R-CNN)}&(W$/$A) &(\%)&(\%)& (\%)& (\%)& (\%)& (\%)\\
		\midrule
		Full-precision &32$/$32&{23.2}&{41.2}&{23.4}&{8.6}&{23.2}&{39.6}  &&& Full-precision &32$/$32&{26.0}&{44.8}&{27.2}&{10.0}&{28.9}&{39.7}\\
		\cmidrule(r){1-18}
		BNN~\cite{nipsHubaraCSEB16}&1$/$1&6.2&15.9&3.8&2.4&10.0&9.9&&&BNN~\cite{nipsHubaraCSEB16}&1$/$1&5.6&14.3&2.6&2.0&8.5&9.3\\
		XNOR-Net~\cite{eccvRastegariORF16}&1$/$1&8.1&19.5&5.6&2.6&8.3&13.3&&&XNOR-Net~\cite{eccvRastegariORF16}&1$/$1&10.4&21.6&8.8&2.7&11.8&15.9\\
		Bi-RealNet~\cite{ijcvLiuLWYLC20}&1$/$1&11.2& 26.0 &8.3&3.1&12.0&18.3&&&Bi-RealNet~\cite{ijcvLiuLWYLC20}&1$/$1&14.4&29.0&13.4&3.7&15.4&24.1\\
		BiDet~\cite{cvprWangWL020}&1$/$1&13.2& 28.3&10.5&5.1&14.3&20.5&&&BiDet~\cite{cvprWangWL020}&1$/$1&15.7&31.0&14.4&4.9&16.7&25.4\\
		\cmidrule(r){1-18}
		Sparks (ours) & 0.78$/$1 & 13.4 & 28.6 & 10.6 &5.3 & 14.5&20.8&&&Sparks (ours) & 0.78$/$1 &15.6&30.7&14.0&4.7&16.5&25.1\\
		Sparks (ours) & 0.56$/$1 & 12.5&27.7&10.0&4.9&14.1& 19.6&&&Sparks (ours) & 0.56$/$1 & 14.9&29.9&13.6&4.1&15.7&24.5  \\
		\cmidrule(r){1-18}
		Sparks$^*$ (ours) & 0.78$/$1 & 18.6 & 35.7 & 17.4 &7.1 & 19.3&31.0&&&Sparks$^*$ (ours) & 0.78$/$1 &21.2&37.5&18.2&7.8&22.6&31.7\\
		Sparks$^*$ (ours) & 0.56$/$1 & 17.6&33.9&17.0&6.6&18.1& 29.4&&&Sparks$^*$ (ours) & 0.56$/$1 & 20.0&36.8&17.4&7.0&20.2&30.5  \\
		\bottomrule
	\end{tabular}}
\end{small}
\end{center}
\vskip-1em
\end{table*}

\subsection{Results}
\label{sec:detection_results}
\textbf{Evaluation on PASCAL VOC.} 
We contrast Sparks against SOTA detection binarization methods including the standard BNN~\cite{nipsHubaraCSEB16}, Bi-RealNet~\cite{ijcvLiuLWYLC20}, XNOR-Net~\cite{eccvRastegariORF16} and BiDet~\cite{cvprWangWL020} in Table \ref{table:object_voc}. We  implement two different versions of Sparks including 0.56-bit and 0.78-bit. Compared with BiDet, our 0.56-bit method obtains about model compression by twice (0.56 vs 1) and computation acceleration by more than 3 times (\emph{e.g.} 285 vs 64 on VGG16+SSD300). 
Besides, by adopting the two-step training scheme and the generalized Sign/PReLU functions, our methods achieve new records on 1-bit object detection.

\textbf{Evaluation on COCO.} 
To further assess the proposed method on a larger and more challenging dataset, we conduct experiments on  COCO. Comparisons with SOTA methods are provided in Table \ref{table:object_coco}. Following the standard COCO evaluation metrics, we report the average mAP over different IoU thresholds from 0.5 to 0.95, the APs at particular thresholds: AP$_{50}$ and AP$_{75}$, and the scale-aware metrics: AP$_{s}$, AP$_{m}$ and AP$_{l}$. 
The benefits of Sparks are still observed, namely, clearly saving in complexity. Results with SSD300 indicate that our 0.78-bit Sparks  even defeats BiDet  in terms of all evaluation metrics. We speculate  that Sparks reduces information redundancy by selecting essential codewords, and thus eliminates some of the false positives. In addition, our method performs stably for both the one-stage SSD300 and the two-stage Faster R-CNN, implying its robustness on different backbones. In addition, results of Sparks$^*$ indicate that our method also has  compatibility with  the two-step training scheme and the generalized functions.

\section{Discussion and Limitation}
In this research, we propose Sparks  that largely enhances both the storage and computation efficiencies of BNNs. Our work is motivated by that kernel-wise codewords are highly clustered. For this reason, we propose a novel selection-based approach for kernel-wise sub-codebook learning instead of previously used channel-wise product quantization. By extending our Sparks with more layers or other blocks, the performance could surpass the standard BNN model with still fewer parameters and BOPs. This provides us with a new research line of training lighter and better BNN models.
As an open-sourced research on well-used benchmarks, our method does not raise ethical concerns. However, one should notice that compressing a model needs to access the model parameters which might need further protection methods for the model privacy.

\vspace{1em}
\begin{multicols}{2}

{
\bibliographystyle{ieee_fullname}
\balance
\bibliography{egbib}

}

\end{multicols}

\end{document}